\documentclass[journal]{IEEEtran}
\IEEEoverridecommandlockouts
\pdfoutput=1

\usepackage{cite}
\usepackage{amsmath,amssymb,amsfonts}
\usepackage{algorithmic}
\usepackage{graphicx}
\usepackage{textcomp}
\usepackage{xcolor}
\usepackage{makecell}
\newcommand{\Mod}[1]{\ (\mathrm{mod}\ #1)}

\def\BibTeX{{\rm B\kern-.05em{\sc i\kern-.025em b}\kern-.08em
    T\kern-.1667em\lower.7ex\hbox{E}\kern-.125emX}}

\usepackage{amsthm}
\newtheorem*{remark}{Remark}

\begin{document}

\title{Context-Aware Ensemble Learning for Time Series}

\author{Arda Fazla, Mustafa E. Aydin, Orhun Tamyigit and Suleyman S. Kozat,~\IEEEmembership{Senior Member,~IEEE}%
\thanks{A. Fazla, M. E. Aydin, O. Tamyigit, S. S. Kozat are with the Deartment of Electrical and Electronics Engineering, Bilkent University, Ankara 06800, Turkey, e-mail: arda@ee.bilkent.edu.tr, enesa@ee.bilkent.edu.tr, tamyigit@ee.bilkent.edu.tr, kozat@ee.bilkent.edu.tr.}}
\maketitle

\begin{abstract}
We investigate ensemble methods for prediction in an online setting. Unlike all the literature in ensembling, for the first time, we introduce a new approach using a meta learner that effectively combines the base model predictions via using a superset of the features that is the union of the base models' feature vectors instead of the predictions themselves. Here, our model does not use the predictions of the base models as inputs to a machine learning algorithm, but choose the best possible combination at each time step based on the state of the problem. We explore three different constraint spaces for the ensembling of the base learners that linearly combines the base predictions, which are \emph{convex} combinations where the components of the ensembling vector are all nonnegative and sum up to 1; \emph{affine} combinations where the weight vector components are required to sum up to 1; and the \emph{unconstrained} combinations where the components are free to take any real value. The constraints are both theoretically analyzed under known statistics and integrated into the learning procedure of the meta learner as a part of the optimization in an automated manner. To show the practical efficiency of the proposed method, we employ a gradient-boosted decision tree and a multi-layer perceptron separately as the meta learners. Our framework is generic so that one can use other machine learning architectures as the ensembler as long as they allow for a custom differentiable loss for minimization. We demonstrate the learning behavior of our algorithm on synthetic data and the significant performance improvements over the conventional methods over various real life datasets, extensively used in the well-known data competitions. Furthermore, we openly share the source code of the proposed method to facilitate further research and comparison.
\end{abstract}

\begin{IEEEkeywords}
ensemble learning, regression, time series, multilayer perceptron (MLP), LightGBM
\end{IEEEkeywords}

\section{Introduction}\label{sec:introduction}
\subsection{Background}\label{sec:background}
\IEEEPARstart{W}{e} study regression (prediction, learning and similar) in time series settings where we receive a data sequence related to a target signal and estimate the signal's next values in an online manner. This problem is extensively studied in the machine learning, computational learning theory and signal processing literatures under different names \cite{sayeed2021deep}, \cite{body_electric}, \cite{gharehbaghi2017deep}. An effective way in such problems as shown in many recent competitions \cite{makridakis2020m4, makridakis2022m5} is using ensemble models for improved performance \cite{ensemble_is_better}, where many base predictor models are first trained separately or successively and then inputted to a merging strategy to get the final prediction. Since there seldom exists a true data generating process in practical time series applications, the ensembling of individual predictors tend to produce better performance than the individual models using diversity, as well known in the machine learning literature \cite{ensemble_is_better_2}. 

Common strategies for ensembling include boosting \cite{gbdt_friedman} and bootstrap aggregation \cite{breiman1996bagging} where the base models are intentionally kept weak either by constraining them by design to underfit or by restricting the amount of data samples and/or features used each base model. The resultant ensembled prediction, however, is a direct average or sum of the predictions of the base models, e.g., taking the mean of the weak predictions in regression, or taking the majority vote in classification. Furthermore, the base models in these ``built-in'' ensemblers are often chosen to be the same predictor in practice, e.g., in random forests \cite{breiman2001random}, gradient-boosted decision trees \cite{gbdt_friedman} and extra-trees \cite{extra_trees}, hard decision trees \cite{cart_friedman} are the base predictors. This not only prevents them from using diverse base regressors but also from providing online prediction as the hard decision trees are not differentiable and therefore require periodical re-fitting against a real-time stream of data, which might be time consuming and/or computationally infeasible.

Apart from these built-in ensembling mechanisms, a ``meta learner'' approach is also used for combining predictions \cite{meta_learner_generic}. In that approach, a separate machine learning model, the meta learner, is trained either over the predictions of the base models or the features extracted from the target sequence, where the output of the meta learner is commonly tailored to mimic a weighted average, i.e., the output vector has its values all nonnegative and sum up to 1; the resultant prediction is then a linear combination of the base predictions. When using the predictions of the base models as the features to the meta machine learning model, however, the diversity of the base predictions should be satisfied as otherwise the correlated features might hinder (as always hinder as we show in our simulations) the learning process of the meta learner \cite{correlated_faetures_hinder_learning}. In practice, it might be hard to find diverse-enough base predictors, which renders this approach less practical \cite{diversity_matters_book}. Moreover, even when using other features as input to the ensemble-learning algorithm in addition to the predictions of the base learners, e.g., those extracted from the time series, the base predictions themselves tend to dominate the other features and the correlation issue re-arises \cite{base_predictions_dominate_others}. Additionally, the ensembling models that do not consider the features the base predictors use tend to produce nonoptimal predictions as they lack the proper context in which these predictions are produced, i.e, some base models can work in certain states better than the others; however, this information is lost in the ensembling process, since it is oblivious to this state information. The weights in such combinations therefore tend to be biased towards the prediction of the ``best'' base predictor, which in turn causes meta learner to overfit to the said base model.

Another important aspect in linear combinations of the predictions is the constraint space of the ensembling weight vector. In the general case, the weight vector is \emph{unconstrained}, i.e., its values can take any real valued number. When the statistics of the combined predictions as well as the target signal are known, naturally, the unconstrained weights achieve the lowest possible error \cite{unconstrained_best}. In practice, however, those statistics are rarely known, and the unsurpassability of the unconstrained combinations is subject to the learning procedures, i.e., whether all the parameters will be correctly learnt or not, and therefore not guaranteed. In fact, since the parameter space is the entire Euclidean space of appropriate dimension, unconstrained combinations might lead to overfitting in ensembling. To this end, we also consider two more combining strategies over the weight vector in order to generate a ``regularization'' effect. These strategies are \emph{affine} combinations where the weight vector is required to have its components sum up to 1 and \emph{convex} combinations which builds on the affine combination by also requiring nonnegativity of the ensembling weights. We emphasize that nondifferentiable ensembler models, e.g., a direct average or boosted trees, are not flexible enough to satisfy the latter two constraints in an automated manner \cite{boosting_isnt_diffable_by_default}.

Here, we effectively combine the base predictions in a context-dependent and base model-agnostic manner. To this end, we propose a machine learning-based meta learner that uses a superset of features in training, which equals or extends the concatenated feature sets of the base models and does not include the predictions themselves as the features to the ensemble-learner. For the linear combination of the base predictions, the ensembler outputs a weight vector which is either unconstrained or amenable to satisfy the affine or convex constraints, all in an online manner. Our motivation is to use the superset feature vector in order to make the combining model to be ``context aware'' and less prone to overfitting to a prominent base model, while exploring various constraint spaces to allow for a regularization effect. In particular, we employ a LightGBM (a gradient boosting machine \cite{ke2017lightgbm}) and a neural network (a multilayer perceptron \cite{hornik1989multilayer}) as the meta learners to show the efficacy of the proposed ensembling approach while emphasizing that the framework is generic enough such that any machine learning model capable of minimizing a custom differentiable loss can be used.

\subsection{Related Work}\label{sec:relatedwork}
Ensemble models are heavily investigated in machine learning, computational learning theory, statistics and signal processing \cite{yang2013effective}, \cite{jacobs1991adaptive}, since they provide superior performance due to using diversity. There are several methods for ensembling, for example, one can train base models independently from each other in parallel and then linearly combine their predictions by a deterministic process, which is called bagging \cite{breiman1996bagging}. In boosting \cite{gbdt_friedman}, base models are trained sequentially, with each model solely focusing on correcting the errors made by the previous ones. On the contrary, the stacking method \cite{jacobs1991adaptive} relies on a separate machine learning model to estimate the linear combination weights of the base model predictions. The main issue with conventional ensembling methods is the process of combining predictions linearly or with some other final meta machine learning algorithm. In most cases, predictions are blended using a simple averaging process, and even if a separate machine learning model is used for learning the combination weights, these weights are learned according to the errors of the base models in the training dataset. Thus, ensemble models are prone to suffer from the underfitting and overfitting issues present in the base models \cite{yang2013effective}. As a remedy, several adjustments to the ensemble methods have been proposed, such as employing different preprocessing techniques for each base model, mixing the training data with independent noise for each model, using a different partition of the training data for each model, etc. \cite{zhang2007neural}. All of these approaches are sub-optimal solutions, as the ensemble learner still only considers the errors of the base models, which as we show in our simulations, is inadequate for the combination weight vector to converge to the optimal weights. The linear combination weights should be produced with respect to the data specific side information vector at each sample to exploit the diversity and avoid the learning difficulties due to high correlations, which we support with our simulations. This is natural since real life time series data often contain different patterns depending on the sample time or some data specific side information vector \cite{xiao2019learning}. For example, a daily sales data, such as the M5 Forecasting Dataset \cite{makridakis2022m5}, may show different patterns for weekdays and weekends. Thus, the linear combination of the base model predictions should be learned according to the day of the week information. A study of the short term forecasting of gas demand in Italy \cite{fabbiani2019ensembling} illustrates that, the daily industrial gas demand consumption shows different patterns depending on the weather temperature, thus, weather temperature should also be fed to the ensemble learner. Therefore, we introduce a scheme where the ensemble model learns to combine base model predictions at each time sample, by learning the relation between the errors of the base models and the data specific side information vector.

There have been previous attempts to linearly combine base model predictions based on a certain side information vector. The second place winners of the M4 Forecasting Competition \cite{montero2020fforma} use the LightGBM model \cite{ke2017lightgbm} as their ensemble model, which learns to produce weights at each sample time, by solely relying on the given side information vector. However, even though the ensemble model predictions rely on the data specific side information vector, the model is still trained based on the errors of the base models in the training dataset. Thus, possible training problems of the base models, such as overfitting, can mislead the ensemble learner to produce nonoptimal combination weights during the test dataset. 

In order to eliminate any training related issues, we introduce a novel training scheme, where the training dataset is partitioned into two distinct sets. The base models are trained on the first partition, and then the ensemble learner is trained by their features, state of the problem and their errors on the second partition of the training dataset. Thus, we alleviate training specific issues by the base models, as these issues are not leaked to the ensemble model. Another problem with the given model is that, each weight assigned to any base model is bound to be nonnegative, and the summation of all of the base model weights for each sample is bound to sum up to $1$, which is the convex constraint. The motive behind this approach is reducing the learning complexity of the ensemble learner. However, as we illustrate in our simulations, the convex constraint is not always the best solution for every time series data. Hence, we solve the linear weight combination problem under three different constraints, where the weights can be either affine constrained, convex constrained or unconstrained. Therefore, we introduce a more comprehensive solution, where the weight constraint employed can be selected according to the given dataset. As a result, with our novel ensemble approach, we significantly improve the prediction performance compared to the base models and conventional ensembling methods for various datasets, as illustrated in our simulations including both well-known datasets in various competitions and artificially generated datasets.

\subsection{Contributions}\label{sec:contributions}
Our contributions are as follows:

\begin{itemize}
    \item For the first time in the literature, we tackle the problem of finding the optimal combination weight vectors to ensemble base predictors based on a sequential data specific side information vector, under three different weight constraints. To learn the optimal time dependent and side-information dependent weights, we present two novel and generic ensembling algorithms, based on boosted decision trees \cite{gbdt_friedman} and neural networks \cite{hornik1989multilayer}, where we derive all the related equations. Note that our approach is generic as any such universal learner can be used accordingly. 
    \item We introduce a novel ensemble training scheme, where we alleviate any training related issues by the base models, such as co-linearity, high correlation and overfitting by the base algorithms.
    \item With various experiments containing synthetic and real-life sequential data from the well-known competitions, we illustrate the superiority of our approach over the base models used in the experiments and the conventional ensembling methods.
    \item We publicly share our code for both model design, comparisons and experimental reproducibility\footnote{https://github.com/ardafazla/context-aware-ensemble}.
\end{itemize}

\subsection{Organization}\label{sec:organization}
The rest of the paper is organized as follows. In Section \ref{sec:preliminaries}, we introduce the problem of finding the optimal base model combination weights based on the side information vector, under different constraints. In Section \ref{sec:theproposedmodel}, we first analyze the associated learning costs of the optimal combination weight vectors, under all three constraints. Next, we introduce two novel and generic machine learning algorithms to find the optimal weights, along with a new ensemble training scheme. We then illustrate the performance of our proposed algorithms via extensive experiments involving real life and synthetic data in Section \ref{sec:sims}. We then conclude our paper with remarks in Section \ref{sec:conclusion}.

\section{Preliminaries}\label{sec:preliminaries}
\subsection{Problem Statement}\label{sec:problemstatement}
In this paper, all vectors are real column vectors and are presented by boldface lowercase letters. Matrices are denoted by boldface uppercase letters. $x^{(k)}$ and $x_{t}^{(k)}$ denotes the $k\textsuperscript{th}$ element of the vectors $\boldsymbol{x}$ and $\boldsymbol{x}_t$, respectively. $\boldsymbol{x}^T$ represents the ordinary transpose of $\boldsymbol{x}$. ${X}_{i, j}$ represents the entry at the $i\textsuperscript{th}$ row and the $j\textsuperscript{th}$ column of the matrix $\boldsymbol{X}$.

We study the online prediction of sequential data using a linear mixture of different learning models. The main sequence we aim to predict is $\{y_t\}$; to this end, we employ $M$ (online) learning models, called base models, each of which uses a side information sequence $\{\boldsymbol{s}_k^{(i)}\}$ for $k \leq t$ and $ i = 1, \ldots, M$. We emphasize that each base model is free to use a different side information vector, e.g., $\{\boldsymbol{s}_k^{(i)}\}$ could include the observed past information $\{y_k\}$ along with model-dependent features. At each time $t$, all the base models produce predictions $\hat{y}_{t}^{(i)}$. We combine these $M$ predictions using a linear scheme such that the ensemble prediction $\hat{y}_{t}^E$ of $y_{t}$ is given as
\begin{equation}\label{eq:linear_comb}
    \hat{y}_{t}^E = \boldsymbol{w}_{t}^T \boldsymbol{\hat{y}}_{t},
\end{equation}
where $\boldsymbol{\hat{y}}_{t} = [\hat{y}_{t}^{(1)}, \ldots, \hat{y}_{t}^{(M)}]^T$ is the base prediction vector of size $M$ consisting of the individual scalar predictions of the base models, and $\boldsymbol{w}_{t} \in \mathbb{R}^M$ is the ensembling weight vector at time $t$. We note that even though the ensembling scheme is linear, the adaptive learning procedure that produces $\boldsymbol{w}_{t}$ could be highly nonlinear as we show in Section \ref{sec:algorithmicdesctription}. Note that other highly nonlinear combination methods including MLPs and decision trees are shown to be highly inadequate in learning the correlated relations as shown in our experiments. When we observe $y_{t}$, we suffer the loss
\begin{align}\label{eq:loss_generic}
    L_t = \ell(y_{t}, \hat{y}_{t}^E) &= \ell(y_{t}, \boldsymbol{w}_{t}^T \boldsymbol{\hat{y}}_{t})\\\nonumber
    &= \ell\big(y_{t}, \sum_{i=1}^M w_{t}^{(i)}\,\hat{y}_{t}^{(i)}\big),
\end{align}
where $\ell$, for example, can be the squared error loss.

\begin{figure}[!t]
\centering
\includegraphics[width=\linewidth]{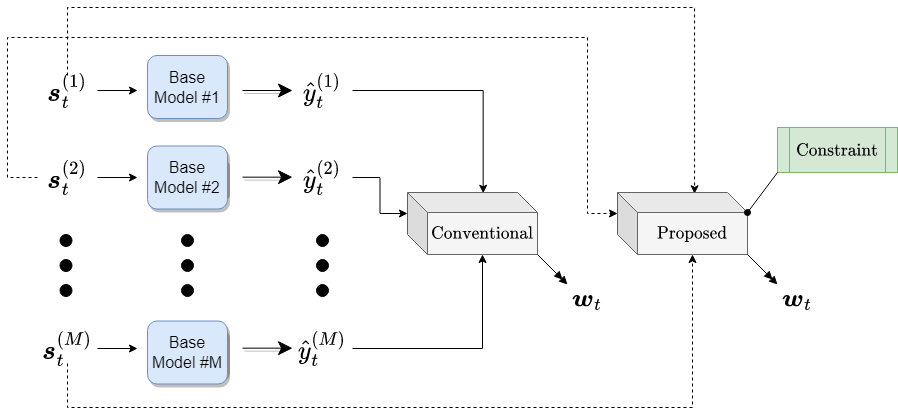}
\caption{Comparison of the ensembling approaches. The conventional approach is shown where the ensembling model is fed with the base predictions only to produce a weight vector at each time $t$. The proposed approach, on the other hand, uses a superset of side information vectors to adaptively learn the combining weights at each time step under a given constraint.}
\label{fig:conventional_vs_new}
\end{figure}

Conventionally, the learning of the weight vector $\boldsymbol{w}_{t}$ depends only on the base models' predictions, i.e., $\boldsymbol{w}_{t} = f(\boldsymbol{\hat{y}}_{t})$ for some function $f$ representing a learning model \cite{conventional_w_learning_1,conventional_w_learning_2}. However, a disadvantage with this approach is that when the base predictions are highly correlated (e.g., due to similar algorithms or outputs), the learning procedure is hindered \cite{correlated_faetures_hinder_learning}. Therefore, highly diverse and independent base models are a prerequisite for such ensembling. Furthermore, this approach does not use any side information in the ensembling model either, which could result in nonoptimal combinations, as, naturally, even with any learning issues,
\begin{equation}\label{eq:side_is_<=}
    \mathbb{E}[\ell(y_{t}, \hat{y}_{t}^E) | \boldsymbol{s}_t^E] \leq \mathbb{E}[\ell(y_{t}, \hat{y}_{t}^E)],
\end{equation}
where $\boldsymbol{s}_t^E$ is the side information vector the ensembling model uses at time step $t$. We note that the loss function $\ell$ in \eqref{eq:side_is_<=} is generic; therefore, \eqref{eq:side_is_<=} implies that the expected loss at time $t$ cannot go higher when using the extra side information vector. To this end, we propose to use \emph{not} the base predictions themselves as inputs to the ensembling model but instead a superset of the side information features of the individual models gathered in $\boldsymbol{s}_t^E$, i.e., it is at least
\begin{equation}\label{eq:superset_side_information}
    \boldsymbol{s}_t^E = \bigcup_{i = 1}^M \boldsymbol{s}_t^{(i)},
\end{equation}
where the union of vectors $\{\boldsymbol{s}_t^{(i)}\}$ corresponds to concatenating of them while not allowing for duplicate features. We note that $\boldsymbol{s}_t^E$ may have more features than this union. Our approach compared to the conventional method is illustrated in Fig. \ref{fig:conventional_vs_new}. We also emphasize that, as shown in Fig. \ref{fig:conventional_vs_new}, we do \emph{not} use the base predictions next to $\boldsymbol{s}_t^E$ as inputs to the ensembling model. The rationale behind this is two folds. Firstly, the base predictions might be highly correlated and as a result, the combining model might struggle in learning the weights \cite{correlated_faetures_hinder_learning}. Secondly, when used along with other features, i.e., $\boldsymbol{s}_t^E$ in this case, the base predictions tend to dominate the other features and the ensemble model would mostly ignore the contributions from possibly valuable side information vectors for ensembling \cite{base_predictions_dominate_others}. In order to emphasize that the learned ensembling weight vector in our framework is side information dependent, we denote it via $\boldsymbol{w}_{\boldsymbol{s}_t^E}$.

As an example of the process, in hourly wind power nowcasting, one could employ two parallel-working base models. The first model could use the past 7 hours of wind power, current wind speed and wind direction, i.e., 3 features in total, while the second model could use past 24 hours of wind power and current wind direction, i.e., 2 features in total, as side information. The ensembling model, then, would use 4 features in total, which are the past 7 and 24 hours of the target signal, wind speed and wind direction. Each base model as well as the ensembling model updates itself as soon as new observations of the next hour become available, i.e., they work in an online manner.

We investigate three different mixture approaches which are identified by how they constrain the combining weight vector $\boldsymbol{w}_{\boldsymbol{s}_t^E}$; these are the \emph{unconstrained} linear combination where each $w_{\boldsymbol{s}_t^E}^{(i)}$ is free to take any real value, the \emph{affine} linear combination where the components of the weight vector sum up to 1, i.e., $\boldsymbol{w}_{\boldsymbol{s}_t^E}^T \boldsymbol{1} = 1$ and the \emph{convex} linear combination where the weight vector not only has its components sum to 1 but also all nonnegative, i.e., $\boldsymbol{w}_{\boldsymbol{s}_t^E}^T \boldsymbol{1} = 1$ and $w_{\boldsymbol{s}_t^E}^{(i)} \geq 0 \,\forall i$; all of these constraints apply to all time steps.

Hence, in a purely online setting, we aim to adaptively find the ``best'' weights that linearly combine the predictions of $M$ parallel running base models to minimize a given loss $\ell$ at each time step $t$ using the super feature set $\boldsymbol{s}_t^E$ under three different constraints imposed on the weight vector. In the next section, we present an analytical overview on the constraints as well as various ensemble algorithms to find the corresponding ensembling weights.

\section{Ensemble Learning}\label{sec:theproposedmodel}
In this section, we first analyze the optimal ensembling vectors in the expectation sense as well as the associated costs for all three constraints on the weight vector; namely, unconstrained, affine and convex combinations. Next, we introduce novel ensembling machine learning algorithms including the mathematical derivations based on boosted decision trees \cite{gbdt_friedman} and neural networks \cite{hornik1989multilayer} in a generic framework to find the optimal weights.

\subsection{Analysis on Constraints}
We first analyze the optimal weight vectors for ensembling as well as the associated costs under known statistics for each of the three constraints imposed on the ensembling weight vector $\boldsymbol{w}_{\boldsymbol{s}_t^E} \in \mathbb{R}^M$ where $M$ is the number of base models in the ensemble. Let $\boldsymbol{C}(\boldsymbol{s}_t)$ be the conditional auto correlation matrix at time $t$ of the base prediction vector $\boldsymbol{\hat{y}}_{t} = [\hat{y}_{t}^{(1)}, \ldots, \hat{y}_{t}^{(M)}]^T$ given the super set of the side information vectors used by all the base algorithms and specific to the ensemble, and let $\boldsymbol{a}(\boldsymbol{s}_t)$ be the conditional cross correlation vector at time $t$ between the target signal $y_{t}$ and the base prediction vector given the super set side information, i.e.,
\begin{align}
    \boldsymbol{C}(\boldsymbol{s}_t) &= \mathbb{E}[\boldsymbol{\hat{y}}_{t} \boldsymbol{\hat{y}}_{t}^T\,|\,\boldsymbol{s}_t^E]\\
    \boldsymbol{a}(\boldsymbol{s}_t) &= \mathbb{E}[{y}_{t} \boldsymbol{\hat{y}}_{t}\,|\,\boldsymbol{s}_t^E].
\end{align}

In the unconstrained case, the components of $\boldsymbol{w}_{\boldsymbol{s}_t^E}$ are free to take any real number; therefore, the corresponding optimization problem given $\boldsymbol{s}_t^E$ for a given differentiable loss function $\ell$ is
\begin{equation}
\begin{aligned}
\min_{\boldsymbol{w}_{\boldsymbol{s}_t^E} \in \mathbb{R}^M} \quad & \ell\big(y_{t}, \sum_{i=1}^M w_{\boldsymbol{s}_t^E}^{(i)}\,\hat{y}_{t}^{(i)}\big),
\end{aligned}
\end{equation}
where the feasible region is the entire $M$ dimensional Euclidean space. If the statistics $\boldsymbol{C}(\boldsymbol{s}_t)$ and 
$\boldsymbol{a}(\boldsymbol{s}_t)$ are known at all times, the optimal unconstrained ensembling vector at time $t$ under the expected squared error loss, i.e., when $\ell(y_t, \hat{y}_t) = \mathbb{E}[(y_t - \hat{y}_t)^2 | \boldsymbol{s}_t^E] = \mathbb{E}[(y_t - \sum_{i=1}^M w_{\boldsymbol{s}_t^E}^{(i)}\,\hat{y}_{t}^{(i)})^2 | \boldsymbol{s}_t^E]$, is the solution to the normal equations and given as
\begin{align}\label{eq:weight_eq_1}
    \boldsymbol{w}_{\boldsymbol{s}_t^E, \text{unc}}^* = \boldsymbol{C}(\boldsymbol{s}_t)^{-1} \boldsymbol{a}(\boldsymbol{s}_t).
\end{align}
Furthermore, assuming the target signal has the conditional variance $\sigma(\boldsymbol{s}_t)^2$ at time $t$, the conditional mean squared error of the unconstrained ensembling is given by
\begin{align}\label{eq:weight_eq_2}
    L_{t, \text{unc}}^* := \sigma(\boldsymbol{s}_t)^2 - \boldsymbol{a}(\boldsymbol{s}_t)^T\boldsymbol{C}(\boldsymbol{s}_t)^{-1} \boldsymbol{a}(\boldsymbol{s}_t).
\end{align}

In the affine constrained case, the components of the weight vector are required to sum up to 1. This translates to the optimization problem given as
\begin{equation}\label{eq:aff_orig}
\begin{aligned}
\min_{\boldsymbol{w}_{\boldsymbol{s}_t^E} \in \mathbb{R}^M} \quad & \ell\big(y_{t}, \sum_{i=1}^M w_{\boldsymbol{s}_t^E}^{(i)}\,\hat{y}_{t}^{(i)}\big)\\
\textrm{subject to} \quad & \boldsymbol{w}_{\boldsymbol{s}_t^E}^T \boldsymbol{1} = 1,
\end{aligned}
\end{equation}
where the feasible region is now an $M-1$ dimensional hyperplane in $\mathbb{R}^M$. In fact, this problem could be cast as an $M-1$-dimensional unconstrained optimization over $\boldsymbol{\tilde{w}}_{\boldsymbol{s}_t} \in \mathbb{R}^{M-1}$ such that the $M^{\textsuperscript{th}}$ component of the weight vector complements the sum of the entire vector to be 1, i.e.,
\begin{equation}\label{eq:aff_trans}
\begin{aligned}
\min_{\boldsymbol{\tilde{w}}_{\boldsymbol{s}_t^E} \in \mathbb{R}^{M-1}} \quad & \ell\big(y_{t}, \sum_{i=1}^{M-1} w_{\boldsymbol{s}_t^E}^{(i)}\,\hat{y}_{t}^{(i)}\big)\\
\textrm{subject to} \quad & w_{\boldsymbol{s}_t^E}^{(M)} = 1 - \boldsymbol{\tilde{w}}_{\boldsymbol{s}_t^E}^T \boldsymbol{1}.
\end{aligned}
\end{equation}
With the transformation of the constrained problem from \eqref{eq:aff_orig}, which is a linearly constrained quadratic optimization problem under the squared loss, i.e., $\ell(y_t, \hat{y}_t) = \mathbb{E}[(y_t - \hat{y}_t)^2 | \boldsymbol{s}_t^E]$, to \eqref{eq:aff_trans}, which is an unconstrained problem, the optimal affine ensembling weight vector admits a closed form solution given as
\begin{align*}
    \boldsymbol{w}_{\boldsymbol{s}_t^E, \text{aff}}^* =  \boldsymbol{C}(\boldsymbol{s}_t)^{-1} \boldsymbol{a}(\boldsymbol{s}_t) - \frac{\boldsymbol{1}^T\boldsymbol{C}(\boldsymbol{s}_t)^{-1} \boldsymbol{a}(\boldsymbol{s}_t) - 1}{\boldsymbol{1}^T\boldsymbol{C}(\boldsymbol{s}_t)^{-1}\boldsymbol{1}}\boldsymbol{C}(\boldsymbol{s}_t)^{-1}\boldsymbol{1}.
\end{align*}
We note that the affine combination preserves the unbiasedness of the base predictors. If each base predictor provides unbiased predictions of $y_{t}$, then we have
\begin{align*}
    \mathbb{E}[\boldsymbol{w}_{\boldsymbol{s}_t^E}^T \boldsymbol{\hat{y}_{t}}] &= \sum_{i=1}^M w_{\boldsymbol{s}_t^E}^{(i)} \mathbb{E}[\hat y_{t}^{(i)}]
    = \sum_{i=1}^M w_{\boldsymbol{s}_t^E}^{(i)} \mathbb{E}[y_{t}]\\
    &= \mathbb{E}[y_{t}] \sum_{i=1}^M w_{\boldsymbol{s}_t^E}^{(i)}
    = \mathbb{E}[y_{t}]. 
\end{align*}
For the expected squared error loss, the optimal affine-constrained weights is given as
\begin{align}
    L_{t, \text{aff}}^* := \sigma(\boldsymbol{s}_t)^2 &- \boldsymbol{a}(\boldsymbol{s}_t)^T\boldsymbol{C}(\boldsymbol{s}_t)^{-1} \boldsymbol{a}(\boldsymbol{s}_t)\\ &+ \frac{(\boldsymbol{1}^T\boldsymbol{C}(\boldsymbol{s}_t)^{-1} \boldsymbol{a}(\boldsymbol{s}_t) - 1)^2}{\boldsymbol{1}^T\boldsymbol{C}(\boldsymbol{s}_t)^{-1}\boldsymbol{1}}.
\end{align}

As for the convex constraint case, we not only require that the components of the weight vector sum up to 1 but also they should be nonnegative, i.e., the optimization problem becomes
\begin{equation}\label{eq:convex_prob}
\begin{aligned}
\min_{\boldsymbol{w}_{\boldsymbol{s}_t^E} \in \mathbb{R}^M} \quad & \ell\big(y_{t}, \sum_{i=1}^M w_{\boldsymbol{s}_t^E}^{(i)}\,\hat{y}_{t}^{(i)}\big)\\
\textrm{subject to} \quad & \boldsymbol{w}_{\boldsymbol{s}_t^E}^T \boldsymbol{1} = 1\\
  &w_{\boldsymbol{s}_t^E}^{(i)}\geq0, \quad i = 1, \ldots, M.    \\
\end{aligned}
\end{equation}
Under a convex loss $\ell$, problem \eqref{eq:convex_prob} is a convex quadratic minimization problem, and the feasible region is the $M$-dimensional unit simplex. Unlike the previous two constraints, \eqref{eq:convex_prob} does not admit a closed form solution. We can, however, project (unconstrained) weights to the unit simplex iteratively to find the optimal weight vector. For brevity, we let $M = 2$, which is a common case in practice, and derive the procedure under squared error loss. To this end, we begin by noting that the squared loss at time $t$ as a function of the ensembling weight vector $\boldsymbol{w}_{\boldsymbol{s}_t^E}$ is
\begin{align}\label{eq:generic_mse}
    &L_t(\boldsymbol{w}_{\boldsymbol{s}_t^E}) =\sigma(\boldsymbol{s}_t)^2 - \boldsymbol{a}(\boldsymbol{s}_t)^T\boldsymbol{C}(\boldsymbol{s}_t)^{-1} \boldsymbol{a}(\boldsymbol{s}_t)\\\nonumber
    &\hspace{0.6cm}+(\boldsymbol{w}_{\boldsymbol{s}_t^E} - \boldsymbol{C}(\boldsymbol{s}_t)^{-1} \boldsymbol{a}(\boldsymbol{s}_t))^T \boldsymbol{C}(\boldsymbol{s}_t)(\boldsymbol{w}_{\boldsymbol{s}_t^E} - \boldsymbol{C}(\boldsymbol{s}_t)^{-1} \boldsymbol{a}(\boldsymbol{s}_t))\\
    &\hspace{1.21cm}= L_{t, unc}^* + (\boldsymbol{w}_{\boldsymbol{s}_t^E} - \boldsymbol{w}_{\boldsymbol{s}_t^E, \text{unc}}^*)^T\boldsymbol{C}(\boldsymbol{s}_t)(\boldsymbol{w}_{\boldsymbol{s}_t^E} - \boldsymbol{w}_{\boldsymbol{s}_t^E, \text{unc}}^*).\nonumber
\end{align}
We can rewrite \eqref{eq:generic_mse} instead of the optimal affine weights $\boldsymbol{w}_{\boldsymbol{s}_t^E, \text{aff}}^*$ as
\begin{align*}
    L_t(\boldsymbol{w}_{\boldsymbol{s}_t^E}) &= L_{t, \text{unc}}^*\\ &\hspace{0.3cm}+((\boldsymbol{w}_{\boldsymbol{s}_t^E} - \boldsymbol{w}_{\boldsymbol{s}_t^E, \text{aff}}^*) + (\boldsymbol{w}_{\boldsymbol{s}_t^E, \text{aff}}^* - \boldsymbol{w}_{\boldsymbol{s}_t^E, \text{unc}}^*))^T\boldsymbol{C}(\boldsymbol{s}_t)\\
    &\hspace{0.9cm}((\boldsymbol{w}_{\boldsymbol{s}_t^E} - \boldsymbol{w}_{\boldsymbol{s}_t^E, \text{aff}}^*) + (\boldsymbol{w}_{\boldsymbol{s}_t^E, \text{aff}}^* - \boldsymbol{w}_{\boldsymbol{s}_t^E, \text{unc}}^*))\\
    &=L_{t, unc}^*\\
&\hspace{0.4cm}+(\boldsymbol{w}_{\boldsymbol{s}_t^E, \text{aff}}^*-\boldsymbol{w}_{\boldsymbol{s}_t^E, \text{unc}}^*)^T\boldsymbol{C}(\boldsymbol{s}_t)(\boldsymbol{w}_{\boldsymbol{s}_t^E, \text{aff}}^*-\boldsymbol{w}_{\boldsymbol{s}_t^E, \text{unc}}^*)\\
&\hspace{0.4cm}+(\boldsymbol{w}_{\boldsymbol{s}_t^E}-\boldsymbol{w}_{\boldsymbol{s}_t^E, \text{aff}}^*)^T\boldsymbol{C}(\boldsymbol{s}_t)(\boldsymbol{w}_{\boldsymbol{s}_t^E}-\boldsymbol{w}_{\boldsymbol{s}_t^E, \text{aff}}^*)\\
&\hspace{0.4cm}-2(\boldsymbol{w}_{\boldsymbol{s}_t^E}-\boldsymbol{w}_{\boldsymbol{s}_t^E, \text{aff}}^*)^T\boldsymbol{C}(\boldsymbol{s}_t)(\boldsymbol{w}_{\boldsymbol{s}_t^E, \text{aff}}^*-\boldsymbol{w}_{\boldsymbol{s}_t^E, \text{unc}}^*)\\
&= L_{t, \text{aff}}^* + (\boldsymbol{w}_{\boldsymbol{s}_t^E}-\boldsymbol{w}_{\boldsymbol{s}_t^E, \text{aff}}^*)^T\boldsymbol{C}(\boldsymbol{s}_t)(\boldsymbol{w}_{\boldsymbol{s}_t^E}-\boldsymbol{w}_{\boldsymbol{s}_t^E, \text{aff}}^*)\\ &\hspace{1.2cm}-2\frac{(\boldsymbol{1}^T\boldsymbol{w}_{\boldsymbol{s}_t^E, \text{unc}}^* - 1)}{\boldsymbol{1}^T\boldsymbol{C}(\boldsymbol{s}_t)^{-1}\boldsymbol{1}}(\boldsymbol{w}_{\boldsymbol{s}_t^E}-\boldsymbol{w}_{\boldsymbol{s}_t^E, \text{aff}}^*)^T\boldsymbol{1}\\
&=L_{t, \text{aff}}^* + (\boldsymbol{w}_{\boldsymbol{s}_t^E}-\boldsymbol{w}_{\boldsymbol{s}_t^E, \text{aff}}^*)^T\boldsymbol{C}(\boldsymbol{s}_t)(\boldsymbol{w}_{\boldsymbol{s}_t^E}-\boldsymbol{w}_{\boldsymbol{s}_t^E, \text{aff}}^*),
\end{align*}
where the last line follows from $(\boldsymbol{w}_{\boldsymbol{s}_t^E}-\boldsymbol{w}_{\boldsymbol{s}_t^E, \text{aff}}^*)^T\boldsymbol{1} = 0$ due to the unit simplex constraint. Further ignoring the constant $L_{t, \text{aff}}^*$, the optimization problem reduces to minimizing the $\boldsymbol{C}(\boldsymbol{s}_t)$-weighted $\ell_2$-norm of $\boldsymbol{w}_{\boldsymbol{s}_t^E}-\boldsymbol{w}_{\boldsymbol{s}_t^E, \text{aff}}^*$ subject to $\boldsymbol{w}_{\boldsymbol{s}_t^E}$ being on the unit simplex. For the assumed $M = 2$, we let $\Delta_2$ be the unit simplex in $\mathbb{R}^2$. There are 2 cases to consider for this problem:
\begin{itemize}
    \item $\boldsymbol{w}_{\boldsymbol{s}_t^E, \text{aff}}^* \in \Delta_{2}$:
    Optimal affine combination is within the unit simplex set and hence a closed form solution can be written for the optimal weight vector: $\boldsymbol{w}_{\boldsymbol{s}_t^E, \text{con}}^* = \boldsymbol{w}_{\boldsymbol{s}_t^E, \text{aff}}^*$. Accordingly, $L_{t, \text{con}}^* := L_{t, \text{aff}}^*$.
    \item $\boldsymbol{w}_{\boldsymbol{s}_t^E, \text{aff}}^* \notin \Delta_{2}$:
    This case is when combination weights are of opposite sign. For simplicity, we assume the weight that is assigned to the first base predictor is negative and the other one is positive. Then, we can write $\boldsymbol{w}_{\boldsymbol{s}_t^E, \text{con}}^* = \boldsymbol{w}_{\boldsymbol{s}_t^E, \text{aff}}^* + \alpha[1\:\:\:-1]^T$ for some $\alpha \in \mathbb{R}$. Consequently, the cost function to minimize becomes the $\boldsymbol{C}(\boldsymbol{s}_t)$-weighted $\ell_2$-norm of $\alpha[-1\:\:\:1]^T$. This implies that the cost is proportional to the magnitude of $\alpha$. The optimal choice would be the smallest value that keeps $\boldsymbol{w}_{\boldsymbol{s}_t^E, \text{con}}^*$ within $\Delta_{2}$, which is $-\,w_{\boldsymbol{s}_t^E, \text{aff}}^{(1)}$. In this case, $\boldsymbol{w}_{\boldsymbol{s}_t^E, \text{con}}^*$ is the corner point of $\Delta_{2}$: $[0\:\:\:1]^T$. This leads to $L_{t, \text{con}}^* := L_{t, \text{aff}}^* + [-1\:\:\:1]\,\boldsymbol{C}(\boldsymbol{s}_t)\,[-1\:\:\:1]^T(w_{\boldsymbol{s}_t^E, \text{aff}}^{(1)})^2.$
\end{itemize}

\begin{remark}
From the inclusion order of the feasible regions of the constrained problems as well as the derived optimal squared error losses as the special cases, we naturally have
\begin{equation}\label{eq:loss_order}
    L_{t, \text{unc}}^* \leq L_{t, \text{aff}}^* \leq L_{t, \text{con}}^*
\end{equation}
for all times $t$. This inequality holds in theory when the correlation statistics of the target signal and the base models are known at all times. In practice, however, those statistics are rarely known and therefore there is a learning cost associated with each constraint, which renders the relation in \eqref{eq:loss_order} unuseful. Hence, even though the unconstrained case seems to promise the least error, the associated learning procedure might lead to overfitting, i.e., in essence, affine and convex constrained cases perform as ``regulators'' in that they offer a tradeoff between bias and variance of the ensembling model. In fact, as shown in our simulations in Section \ref{sec:sims}, the unconstrained ensembling scheme does not always achieve the lowest error.
\end{remark}

Since $\boldsymbol{C}(\boldsymbol{s}_t)$ and $\boldsymbol{a}(\boldsymbol{s}_t)$ are rarely known in practice, next, we present novel and generic ensembling algorithms to find the optimal ensembling weight vectors under three constraint schemes.

\begin{figure*}[!t]
\centering
\includegraphics[width=0.8\linewidth]{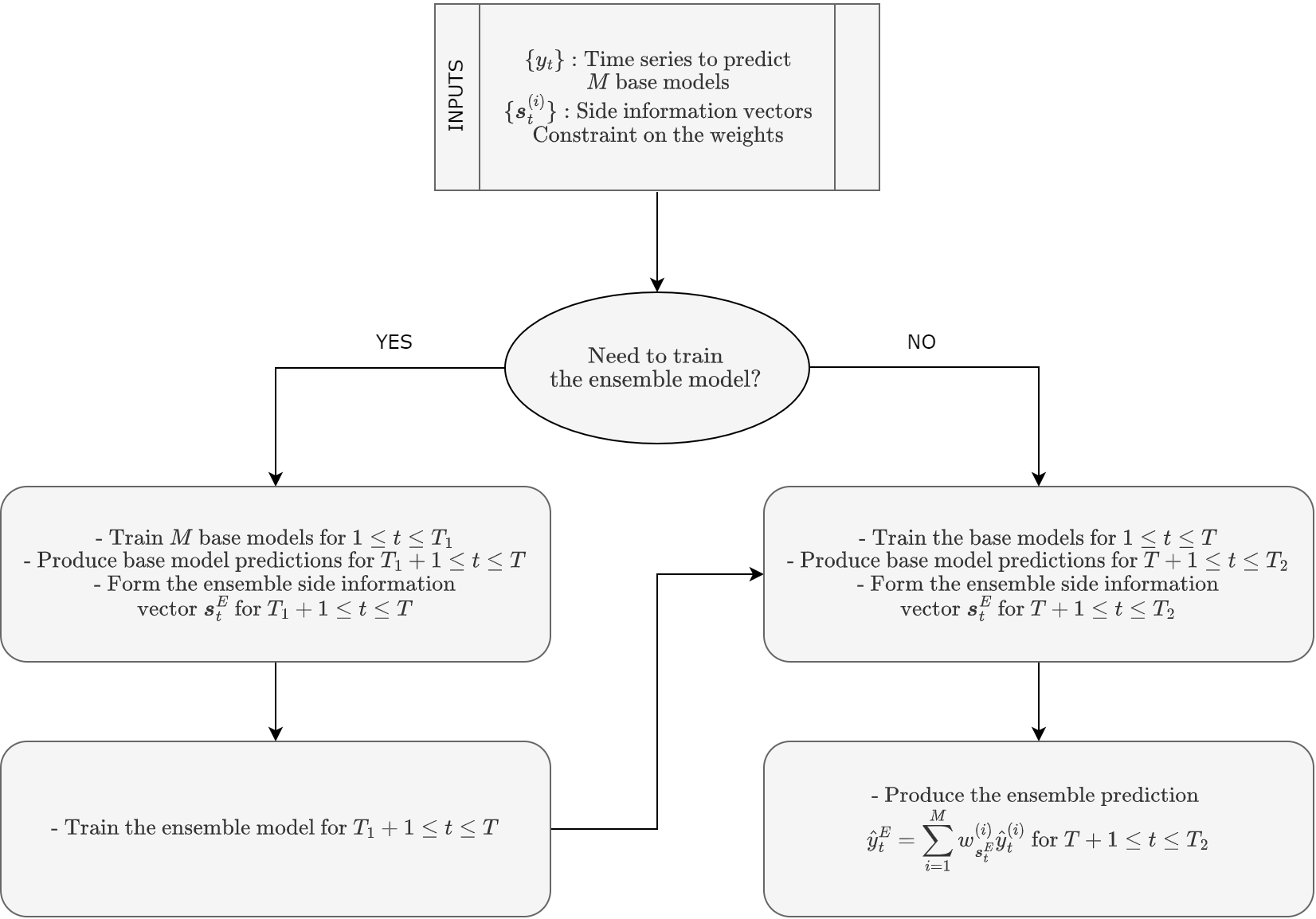}
\caption{Flowchart of the proposed ensemble training scheme.}
\label{fig:alg_flow}
\end{figure*}

\subsection{Introduced Approach}\label{sec:algorithmicdesctription}

In order to linearly combine the predictions of different base models by assigning weights to each model, we use ensembling algorithms. Our ensemble algorithms learn to combine the predictions of different base algorithms, by minimizing the loss function given in \eqref{eq:loss_generic}, where $y_{t}$ is the observed sequence at time $t$, $\ell$ is a given differentiable loss function such as the squared error loss, $\hat y^{E}_{t}$ is the prediction of the ensemble algorithm, $M$ is the number of base models, $\hat y^{(i)}_{t}$ are the predictions of the base models and $w_{\boldsymbol{s}^E_{t}}^{(i)}$ are the weights assigned to the base models, by utilizing the side information sequence $\boldsymbol{s}^E_{t}$.

In order to train the ensemble algorithm, we introduce a novel ensemble training scheme, illustrated in Fig. \ref{fig:alg_flow} and explained thoroughly in \textbf{Algorithm 1}. Our training scheme consists of two phases: the offline phase and the online phase. In the offline phase, first, the base models are independently trained on a certain partition of the training dataset, given as $1 \leq t \leq T_{1}$. Then, the predictions of the base models are gathered on the remaining partition of the training data, given as $T_{1}+1 \leq t \leq T$, where these predictions are then used to train the ensemble algorithm. Hence, as the predictions are gathered on unseen data, we overcome the possible overfitting issue of the base models, which is a well-known issue for the conventional ensembling approaches \cite{yang2013effective}. Each base model $i$ makes predictions based on a side information vector ${\boldsymbol{s}_{t}^{(i)}}$, which may consist of the past observations of the sequence $y_{t}$ along with some additional related information specific to the observed sequence or the base model. Our ensemble model is trained with the errors of the base models, while employing the side information sequence $\boldsymbol{s}^E_{t}$, which may consist of the past observations of the sequence $y_{t}$, side information related to base models, e.g., errors of the base models, and some additional information specific to the observed sequence. The side information vector of the ensemble model is created as explained in \eqref{eq:superset_side_information}.

In the online phase, we use our pre-trained ensemble algorithm from the offline phase to produce combination weights for the base algorithms on the new data, which in our case, is the new observations following the training samples of the sequence $y_{t}$. First, we retrain the base models on the whole training dataset, given as $1 \leq t \leq T$. Then, we produce the base model predictions for the test dataset, $T+1 \leq t \leq T_{2}$. We then update the ensemble side information vector $\boldsymbol{s}^E_{t}$ for the test duration. Finally, we produce the ensemble prediction $\hat y^{E}_{t} = \sum_{i=1} ^{M} w_{\boldsymbol{s}^E_{t}}^{(i)} \hat y^{{(i)}}_{t}$, for $T+1 \leq t \leq T_{2}$. \\
---------------------------------------------------------------------------\\
\textbf{Algorithm 1} Ensemble Learning Procedure \\
---------------------------------------------------------------------------
\begin{itemize}

    \item \textbf{Offline Phase:} Training the Base and Ensemble Algorithms
    \\\textbf{inputs}:
    \begin{itemize}
        \item ${\{{y_t}\}}\triangleq{\{{y_t}\}}_{t=1}^{T}$: the training data that consists of $T$ sequential samples from the time series sequence $y_{t}$.
        \item $M$ base models.
        \item ${\{\boldsymbol{s}_{t}^{(i)}\}}\triangleq{\{\boldsymbol{s}_{t}^{(i)}\}}_{t=1}^{T}$: time series features for the $i\textsuperscript{th}$ base predictor at time $T$.
    \end{itemize}
    \textbf{outputs}:
    \begin{itemize}
    \item ensemble learner, which produces the weight vector $w_{\boldsymbol{s}^E_{t}}$ for combining base predictions.
    \item ${\{\boldsymbol{s}_{t}^{E}\}}\triangleq{\{\boldsymbol{s}_{t}^{E}\}}_{t=1}^{T}$: time series features for the ensemble learner.
    \end{itemize}
    \textit{\textbf{prepare ensemble data}}:
    \begin{itemize}
    \item split ${\{{y_t}\}}\triangleq{\{{y_t}\}}_{t=1}^{T}$ into training and test periods for the base algorithms, where $1 \leq t \leq T_{1}$ is the training period and $T_{1} + 1 \leq t \leq T$ is the test period.
    \item train all of the base algorithms over the training period.
    \item generate forecasts for the test period for all of the base models, i.e., form ${\{{\hat{y}_t}^{(i)}\}}_{t= T_{1}+1}^{T},\,\, {1 \leq i \leq M} $.
    \item construct the ensemble side information vector ${\boldsymbol{s}_{t}^{E}}$ by following the procedure in \eqref{eq:superset_side_information}.
    \end{itemize}
    \textit{\textbf{train the ensemble learner}}:
    \begin{itemize}
        \item train the ensemble learner using ${\boldsymbol{s}_{t}^{E}}$ to minimize the loss
        \begin{equation}
        \label{eq:lgbm_loss}
            \arg \min_{\boldsymbol{w}^T_{\boldsymbol{s}^E_{t}}} \sum_{t=T_{1}+1}^{T} \ell(y_{t} , \hat y^{E}_{t})
        \end{equation}
        where $\hat y^{E}_{t} = \sum_{i=1} ^{M} w_{\boldsymbol{s}^E_{t}}^{(i)} \hat y^{(i)}_{t}$, $\ell$ is a given differentiable loss function such as the squared error loss, $\boldsymbol{w}_{\boldsymbol{s}^E_{t}} = [\,{w_{\boldsymbol{s}^E_{t}}^{(1)}},\,{w_{\boldsymbol{s}^E_{t}}^{(2)}},\,\dots,\,{w_{\boldsymbol{s}^E_{t}}^{(M)}}\,]^T$ is the weight vector for the linear combination of the base learners, and $w_{\boldsymbol{s}^E_{t}}^{(i)}$ is the combination weight assigned to the $i\textsuperscript{th}$ base learner at time $t$.\\
        \item train all of the base algorithms over the entire training data ${1 \leq t \leq T}$.
    \end{itemize}
    \item \textbf{Online Phase:} Predicting the Test Data
    \\\textbf{inputs}:
    \begin{itemize}
        \item ${\{{y_t}\}}_{t=T+1}^{T_{2}}$: the test data that consists of sequential data samples of the time series sequence $y_{t}$, from $T+1$ to $T_{2}$.
        \item the trained ensemble learner from the offline phase.
        \item $M$ base models, trained on the training data.
        \item ${\{\boldsymbol{s}_{t}^{(i)}\}}_{t=T+1}^{T_{2}}$: time series features for the $i\textsuperscript{th}$ base predictor at time $T$.
        \item ${\{\boldsymbol{s}_{t}^{E}\}}_{t=T+1}^{T_{2}}$: time series features for the ensemble learner.
    \end{itemize}
    \textbf{output}:
    \begin{itemize}
        \item combination weights $w_{\boldsymbol{s}^E_{t}}^{(i)}$ for each base learner ${1 \leq i \leq M}$, for ${T+1 \leq t \leq T_{2}}$.
        \item ensemble predictions $\hat y^{E}_{t}$ for ${T+1 \leq t \leq T_{2}}$.
    \end{itemize}
    \textit{\textbf{generate combination weights}}:
    \begin{itemize}
        \item generate forecasts for the test data from all of the base models, i.e., generate ${\{{\hat{y}_t}^{(i)}\}}_{t= T+1}^{T_{2}},\,\, {1 \leq i \leq M} $.
        \item construct the ensemble side information vector ${\boldsymbol{s}_{t}^{E}}$ for the test dataset by following the procedure in \eqref{eq:superset_side_information}.
        \item feed the side information vector ${\boldsymbol{s}_{t}^{E}}$ to the ensemble learner to generate the weight vector $\boldsymbol{w}^T_{\boldsymbol{s}^E_{t}} = [\,{w_{\boldsymbol{s}^E_{t}}^{(1)}},\,{w_{\boldsymbol{s}^E_{t}}^{(2)}},\,\dots,\,{w_{\boldsymbol{s}^E_{t}}^{(M)}}\,]^T$, for $T+1 \leq t \leq T_{2}$.
        \item linearly combine the weights given to each base learner with corresponding predictions to predict $\hat y^{E}_{t} = \sum_{i=1} ^{M} w_{\boldsymbol{s}^E_{t}}^{(i)} \hat y^{{(i)}}_{t}$, for $T+1 \leq t \leq T_{2}$.
    \end{itemize}
\end{itemize}

We next introduce two novel and generic ensemble algorithms which will achieve the optimal weights under different constraints given in \eqref{eq:weight_eq_1},  \eqref{eq:weight_eq_2} and \eqref{eq:aff_orig}.

\subsection{LightGBM Ensemble}\label{sec:lgbm_ens}
Gradient boosting decision trees (GBDT) are sequentially trained decision tree ensembles. In each iteration, GBDT minimizes the residual error made by the previous trees in the sequence. LightGBM is a state-of-the-art GBDT model that deviates in finding split points to minimize a certain loss function for the whole data during training, and is efficient in terms of memory consumption and training speed compared to its counterparts \cite{ke2017lightgbm}.

We employ LightGBM to search for the optimal ensemble combination weights, as described in \textbf{Algorithm 1}. LightGBM requires the gradient and hessian of the objective function \eqref{eq:loss_generic} in order to minimize the loss with respect to the combination weight vector $\boldsymbol{w}_{\boldsymbol{s}^E_{t}}$. We study the linear combination of the base predictions under unconstrained, convex constrained and affine constrained conditions, which were studied in Section \ref{sec:theproposedmodel}. For each base learner $i$, LightGBM produces an output value $p_{\boldsymbol{s}^E_{t}}^{(i)}$, where we then apply the necessary transformation to achieve $w_{\boldsymbol{s}^E_{t}}^{(i)}$ under the given weight constraint. Therefore, for all three cases, we study and provide the gradient and hessian of \eqref{eq:loss_generic} with respect to $p_{\boldsymbol{s}^E_{t}}^{(i)}$, for each base learner $i$. For all constraints, we set the loss function $\ell$ as the squared error loss, although any differentiable loss function can be selected, and assume we have the base predictions for $1 \leq t \leq T$, as explained in Fig. \ref{fig:alg_flow} and \textbf{Algorithm 1}. Therefore, at any given time $t$ from the training dataset, the loss to minimized by the LightGBM ensemble is
\begin{equation*}
    L_{t} =  \ell(y_{t}, \hat y^{E}_{t}) = (y_{t} - \hat y^{E}_{t})^2 = (y_{t} - \sum_{i=1} ^{M} w_{\boldsymbol{s}^E_{t}}^{(i)} \hat y^{(i)}_{t})^2,
\end{equation*}
where $y_{t}$ is the observed sequence, and $\hat y^{E}_{t}$ is the prediction of the ensemble algorithm, at time $t$. The transformation from $p_{\boldsymbol{s}^E_{t}}^{(i)}$ to $w_{\boldsymbol{s}^E_{t}}^{(i)}$ is given as $w_{\boldsymbol{s}^E_{t}}^{(i)} = \tau(p_{\boldsymbol{s}^E_{t}}^{(i)})$, where $\tau$ indicates the transformation operation and is specific to the given weight constraint. Hence, the gradient and hessian to be calculated at any time $t$, for the base learner $i$, is given as

\begin{equation}
\label{eq:gradient_initial}
    G_{i} = \frac{\partial{L_{t}}}{\partial{p_{\boldsymbol{s}^E_{t}}^{(i)}}}, H_{i} = \frac{\partial{G_{i}}}{\partial{p_{\boldsymbol{s}^E_{t}}^{(i)}}}.
\end{equation}

Next, we explain the necessary transformations, and provide the gradient and hessian calculations for all three constraints.

\subsubsection{Unconstrained Case}\label{lgbm_unconstrained}

For the unconstrained case, there is no restriction on the value of $w_{\boldsymbol{s}^E_{t}}^{(i)}$. Therefore, we can simply write the transformation relation as $w_{\boldsymbol{s}^E_{t}}^{(i)} = p_{\boldsymbol{s}^E_{t}}^{(i)}$. The gradient and hessian equations in \eqref{eq:gradient_initial} for the base learner $i$, at time $t$ become
\begin{align*}
    G_{i} &= 2 (y_{t} - \hat y^{E}_{t}) (-\frac{\partial{\hat y^{E}_{t}}}{\partial{p_{\boldsymbol{s}^E_{t}}^{(i)}}})\\
    &= 2 (y_{t} - \hat y^{E}_{t}) (-\sum_{m=1}^M \frac{\partial w_{\boldsymbol{s}^E_{t}}^{(m)}}{\partial p_{\boldsymbol{s}^E_{t}}^{(i)}}\hat y^{(m)}_{t})\\
    &= 2 \hat y^{(i)}_{t} (\hat y^{E}_{t} - y_{t})\\
    H_{i} &= 2 \hat y^{(i)}_{t} (\frac{\partial{\hat y^{E}_{t}}}{\partial{p_{\boldsymbol{s}^E_{t}}^{(i)}}})\\
    &= 2 (\hat y^{(i)}_{t})^2.
\end{align*}

\subsubsection{Affine Constrained Case}\label{affine_constraint}

For the affine constrained case, the restriction is the summation of the combination weights should be equal to one, i.e., $\sum_{i=1} ^{M} w_{\boldsymbol{s}^E_{t}}^{(i)} = 1$. In order to satisfy this relation, we apply the given normalization operation $w_{\boldsymbol{s}^E_{t}}^{(i)} = \frac{p_{\boldsymbol{s}^E_{t}}^{(i)}}{\sum_{m=1}^M p_{\boldsymbol{s}^E_{t}}^{(m)}}$. Hence, the gradient and hessian equations in \eqref{eq:gradient_initial} for the base learner $i$, at time $t$ become
\begin{align*}
    G_{i} &= 2 (y_{t} - \hat y^{E}_{t}) (-\frac{\partial{\hat y^{E}_{t}}}{\partial{p_{\boldsymbol{s}^E_{t}}^{(i)}}})\\
    &= 2 (y_{t} - \hat y^{E}_{t}) (-\sum_{m=1}^M \frac{\partial w_{\boldsymbol{s}^E_{t}}^{(m)}}{\partial p_{\boldsymbol{s}^E_{t}}^{(i)}}\hat y^{(m)}_{t})\\
    &= 2 (y_{t} - \hat y^{E}_{t}) (\frac{1}{\sum_{m=1}^M p_{\boldsymbol{s}^E_{t}}^{(m)}}) (\hat y^{E}_{t} - \hat y^{(i)}_{t}) \\
    H_{i} &= 2 (\frac{1}{\sum_{m=1}^M p_{\boldsymbol{s}^E_{t}}^{(m)}})^2 (\hat y^{E}_{t} - \hat y^{(i)}_{t}) (3 \hat y^{E}_{t} - 2 y_{t} - \hat y^{(i)}_{t})
\end{align*}

\subsubsection{Convex Constrained Case}\label{convex_constraint}

For the convex constrained case, the restriction is the summation of the combination weights should be equal to one, such that $\sum_{i=1} ^{M} w_{\boldsymbol{s}^E_{t}}^{(i)} = 1$. In addition, $0 \leq w_{\boldsymbol{s}^E_{t}}^{(i)} \leq 1$ should also be satisfied. Therefore, we apply the softmax transformation given as $w_{\boldsymbol{s}^E_{t}}^{(i)} = \frac{e^{p_{\boldsymbol{s}^E_{t}}^{(i)}}}{\sum_{m=1}^M e^{p_{\boldsymbol{s}^E_{t}}^{(m)}}}$. Therefore, the gradient and hessian equations in \eqref{eq:gradient_initial} for the base learner $i$, at time $t$ become
\begin{align*}
    G_{i} &= 2 (y_{t} - \hat y^{E}_{t}) (-\frac{\partial{\hat y^{E}_{t}}}{\partial{p_{\boldsymbol{s}^E_{t}}^{(i)}}})\\
    &= 2 (y_{t} - \hat y^{E}_{t}) (-\sum_{m=1}^M \frac{\partial w_{\boldsymbol{s}^E_{t}}^{(m)}}{\partial p_{\boldsymbol{s}^E_{t}}^{(i)}}\hat y^{(m)}_{t})\\
    &= 2 (y_{t} - \hat y^{E}_{t}) w_{\boldsymbol{s}^E_{t}}^{(i)} (\hat y^{E}_{t} - \hat y^{(i)}_{t}) \\
    H_{i} &= G_{i} (1 - 2w_{\boldsymbol{s}^E_{t}}^{(i)}) + 2(w_{\boldsymbol{s}^E_{t}}^{(i)})^2(\hat y^{E}_{t} - \hat y^{(i)}_{t})^2.
\end{align*}

\subsection{MLP Ensemble}\label{sec:mlp_ens}

Artificial Neural Networks(ANN) are models that are based on the processing structure of the brain cells, and are used to model complex patterns and problems \cite{jain1996artificial}. As our ensemble algorithm, we employ a certain class of ANN called Multilayer Perceptron (MLP), which is an ANN with more than two layers. MLPs are heavily used in machine learning due to their ability of learning complex and non-linear relations with high training speed compared to their counterparts \cite{gardner1998artificial}.

As a specific example, we provide the equations for a two layered network architecture, which consists of an input layer, a hidden layer and an output layer. For more layers, naturally, our derivations can be straightforwardly extended. At the end of the output layer, we apply a transformation operation which maps the weights of the ensemble algorithm according to the given weight constraint. We study the linear combination of the base predictions under affine unconstrained, affine constrained and convex constrained conditions, which were studied in Section \ref{sec:theproposedmodel}. The feed-forward equations of our model is given as
\begin{align*}
    \boldsymbol{v} &= \boldsymbol{U}^{(1)}\boldsymbol{x}\\
    \boldsymbol{z} &= f(\boldsymbol{v})\\
    \boldsymbol{p} &= \boldsymbol{U}^{(2)}\boldsymbol{z}\\
    \boldsymbol{w} &= \tau (\boldsymbol{p}),
\end{align*}
where $\boldsymbol{x} \in \mathbb{R}^K$ is the input vector, $\boldsymbol{U}^{(1)} \in \mathbb{R}^{L\text{x}K}$ are the layer weights connecting the input layer to the hidden layer, $\boldsymbol{z} \in \mathbb{R}^L$ is the input of the hidden layer, which is the rectified linear activation function (ReLU) $f$ applied to $\boldsymbol{v} \in \mathbb{R}^L$. $\boldsymbol{U}^{(2)} \in \mathbb{R}^{K\text{x}M}$ are the layer weights connecting the hidden layer to the output layer, $\boldsymbol{p} \in \mathbb{R}^M$ is the output vector of the model, and $\boldsymbol{w} \in \mathbb{R}^M$ is the weight vector for the linear combination of the base models, obtained by applying the transformation operation $\tau$ to $\boldsymbol{p}$. $K, L, M$ are the lengths of the input, hidden and output layers, respectively.

Our ensemble algorithm learns to produce linear combination weights, by minimizing the loss function \eqref{eq:loss_generic}. For all constraints, we set $\ell$ as the squared error loss, although any differentiable loss function can be selected, and assume we have the base predictions for $1 \leq t \leq T$, as explained in Fig. \ref{fig:alg_flow} and \textbf{Algorithm 1}. Therefore, at any given time $t$ from the training data, the loss to minimized by the MLP ensemble is given as
\begin{equation*}
    L_{t} =  \ell(y_{t}, \hat y^{E}_{t}) = (y_{t} - \hat y^{E}_{t})^2 = (y_{t} - \sum_{i=1} ^{M} w_{\boldsymbol{s}^E_{t}}^{(i)} \hat y^{(i)}_{t})^2,
\end{equation*}
where $y_{t}$ is the observed sequence, and $\hat y^{E}_{t}$ is the prediction of the ensemble algorithm, at time $t$. Here, we employ the notations $x_i \triangleq x_{\boldsymbol{s}^E_{t}}^{(i)}$,
$w_i \triangleq w_{\boldsymbol{s}^E_{t}}^{(i)}$ and $p_i \triangleq p_{\boldsymbol{s}^E_{t}}^{(i)}$, for simplicity. The loss is minimized by iteratively updating the layer weights $\boldsymbol{U}^{(1)}$ and $\boldsymbol{U}^{(2)}$ with backpropagation. The backpropagation equations are given as
\begin{align*}
    \frac{\partial{L_{t}}}{\partial{U^{(1)}_{l,k}}} &= \sum_{j=1}^{M}\sum_{i=1}^{M}
    \frac{\partial{L_{t}}}{\partial{w_i}} \frac{\partial{w_i}}{\partial{p_j}} \frac{\partial{p_j}}{\partial{z_l}}
    \frac{\partial{z_l}}{\partial{v_l}}
    \frac{\partial{v_l}}{\partial{U^{(1)}_{l,k}}},\\
    \frac{\partial{L_{t}}}{\partial{U^{(2)}_{m,l}}} &= \sum_{i=1}^M \frac{\partial{L_{t}}}{\partial{w_i}} \frac{\partial{w_i}}{\partial{p_m}} \frac{\partial{p_m}}{\partial{U^{(2)}_{m,l}}},
\end{align*}
where $1 \leq l \leq L$, $1 \leq k \leq K$, $1 \leq m \leq M$. We then update $\boldsymbol{U}^{(1)}$ and $\boldsymbol{U}^{(2)}$ using stochastic gradient descent (SGD). The update equations are given as
\begin{align*}
    U^{(1)}_{l,k} = U^{(1)}_{l,k} - \alpha \frac{\partial{L_{t}}}{\partial{U^{(1)}_{l,k}}}, \;\;
    U^{(2)}_{m,l} = U^{(2)}_{m,l} - \alpha \frac{\partial{L_{t}}}{\partial{U^{(2)}_{m,l}}},
\end{align*}
where $\alpha$ is the learning rate hyperparameter of SGD. Next, we explain the necessary transformations, and provide the closed-form backpropagation equations under the given constraints.

\subsubsection{Unconstrained Case}\label{mlp_unconstrained}

For the unconstrained case, there is no restriction on the value of $w_{\boldsymbol{s}^E_{t}}^{(i)}$. Therefore, we can simply write the transformation relation as $w_{\boldsymbol{s}^E_{t}}^{(i)} = p_{\boldsymbol{s}^E_{t}}^{(i)}$. Hence, the backpropagation equations become
\begin{align*}
    \frac{\partial{L_{t}}}{\partial{U^{(1)}_{l,k}}} &= \sum_{i=1}^M 2 (\hat y^{E}_{t} - y_{t}) \hat y^{(i)}_{t} U^{(2)}_{m,l} f'(v_l) x_k,\\
    \frac{\partial{L_{t}}}{\partial{U^{(2)}_{m,l}}} &= 2 (\hat y^{E}_{t} - y_{t}) \hat y^{(m)}_{t} z_l,
\end{align*}
where $f'(v_{l})$ is the derivative of ReLU, which is the piece-wise function
\begin{equation*}
    f'(v_{l}) = 
    \begin{cases}
    0, & \text{if } {v_{l} \leq 0} \\
    v_{l}, & \text{if } {v_{l} > 0}.
    \end{cases}
\end{equation*}

\subsubsection{Affine Constrained Case}\label{affine_mlp}

For the affine constrained case, the restriction is the summation of the combination weights should be equal to one, such that $\sum_{i=1} ^{M} w_{\boldsymbol{s}^E_{t}}^{(i)} = 1$. In order to satisfy this relation, we apply the given normalization operation $w_{\boldsymbol{s}^E_{t}}^{(i)} = \frac{p_{\boldsymbol{s}^E_{t}}^{(i)}}{\sum_{m=1}^M p_{\boldsymbol{s}^E_{t}}^{(m)}}$.
Hence, the backpropagation equations become
\begin{align*}
    \frac{\partial{L_{t}}}{\partial{U^{(1)}_{l,k}}} &= \sum_{i=1}^M 2 c (\hat y^{E}_{t} - y_{t})(\hat y^{(m)}_{t} - \hat y^{E}_{t}) U^{(2)}_{m,l} f'(v_l) x_k,\\
    \frac{\partial{L_{t}}}{\partial{U^{(2)}_{m,l}}} &= 2 c (\hat y^{E}_{t} - y_{t})(\hat y^{(m)}_{t} - \hat y^{E}_{t}) z_l,
\end{align*}
where $c$ is the constant term given as $c$ = $(\frac{1}{\sum_{m=1}^M p_{\boldsymbol{s}^E_{t}}^{(m)}})$.

\subsubsection{Convex Constrained Case}\label{mlp_convex}

In the convex constrained case, the restriction is the summation of the combination weights should be equal to one, such that $\sum_{i=1} ^{M} w_{\boldsymbol{s}^E_{t}}^{(i)} = 1$. In addition, $0 \leq w_{\boldsymbol{s}^E_{t}}^{(i)} \leq 1$ should also be satisfied. Therefore, we apply the given softmax transformation $w_{\boldsymbol{s}^E_{t}}^{(i)} = \frac{e^{p_{\boldsymbol{s}^E_{t}}^{(i)}}}{\sum_{m=1}^M e^{p_{\boldsymbol{s}^E_{t}}^{(m)}}}$. Hence, the backpropagation equations become
\begin{align*}
    \frac{\partial{L_{t}}}{\partial{U^{(1)}_{l,k}}} &= \sum_{i=1}^M 2 (\hat y^{E}_{t} - y_{t}) w_m  (\hat y^{(m)}_{t} - \hat y^{E}_{t}) U^{(2)}_{m,l} f'(v_l) x_k,\\
    \frac{\partial{L_{t}}}{\partial{U^{(2)}_{m,l}}} &= 2 (\hat y^{E}_{t} - y_{t}) w_m  (\hat y^{(m)}_{t} - \hat y^{E}_{t}) z_l.
\end{align*}

We now have the update equations to train both the MLP ensemble and LightGBM ensemble algorithms according to Fig. \ref{fig:alg_flow} and \textbf{Algorithm 1}. 

\begin{remark}
Note that, although we provide two different models to use as our ensemble learners, any machine learning model such as random forests \cite{breiman2001random}, linear regression models \cite{seber2012linear} etc. can be employed.
\end{remark}

The following section illustrates the simulations of our ensemble models under different constraints, and also shows their superiority over the base models that we employ.

\section{Simulations}\label{sec:sims}

In this section, we first verify our model by using synthetically generated data to show the learning procedure of our algorithms with respect to the conventional classical ensemble methods widely used in the literature \cite{ren2016ensemble}. Then, we showcase the performance of our ensemble models under different constraints using real-life data, which are the well-known prediction competition M5 \cite{makridakis2022m5}, and the daily energy consumption production in Turkey. As the conventional ensemble models, we employ two different models. First, we use a conventional linear ensemble model where linear regression is used to learn the relation between the base model predictions and the observed sequence $y_{t}$. Next, we use an MLP ensemble with two layers to learn the relation between the base model predictions and the observed sequence $y_{t}$. Thus, both of the conventional models do not employ any side information related to the sequence, and make use of only the base model predictions while learning the combination weights.

For both our MLP ensemble introduced in Section \ref{sec:mlp_ens} and LightGBM ensemble introduced in Section \ref{sec:lgbm_ens}, we use two base models and search for the optimal linear combination weights $w_{\boldsymbol{s}^E_{t}}^{(1)}$ and $w_{\boldsymbol{s}^E_{t}}^{(2)}$, given the side information vector $\boldsymbol{s}^E_{t}$, under unconstrained, affine constrained and convex constrained conditions. However, our results can be straightforwardly extended to cases with more than two base models. 
During all our simulations, the data is divided into two partitions as the training and test datasets according to Fig. \ref{fig:alg_flow} and \textbf{Algorithm 1}. We only consider one-step-ahead forecasting and compare models in terms of the total loss
\begin{equation*}
    \sum_{t=T+1}^{T_{2}}L_{t} = \sum_{t=T+1}^{T_{2}}\ell(y_{t} , \hat y_{t}),
\end{equation*}
where $y_{t}$ is the observed sequence at time $t$, $\ell$ is selected as the squared error loss, $T+1 \leq t \leq T_{2}$ is the duration of the test dataset, $1 \leq t \leq T$ is the duration of the training dataset, and $\hat y_{t}$ is the prediction of the corresponding (base or ensemble) model. We also illustrate the cumulative normalized total error for all of the models used in the experiments, given as
\begin{equation*}
    \sum_{k=T+1}^t \frac{(y_k-\hat{y}_k)^2}{t}, t=T+1,\ldots,T_{2},
\end{equation*}
where ${y}_k$ is the current sample of the signal to be predicted and $\hat{y}_k$ is the prediction of the corresponding model.

\subsection{Synthetic Data}
The synthetic data consists of two manually generated and independent data components. First data is generated using the autoregressive integrated moving average (ARIMA) model \cite{box2015time} given by
\begin{equation*}
    y^{\{1\}}_{t} = 0.2y^{\{1\}}_{t-1} - 0.1y^{\{1\}}_{t-2} + 0.3e^{\{1\}}_{t-1} - 0.1e^{\{1\}}_{t-2} + v^{\{1\}}_{t},
\end{equation*}
where $v^{\{1\}}_{t}$ is a sample function from a stationary white Gaussian process with unit variance and $e^{\{1\}}_{t}$ is the error term given by $(y^{\{1\}}_{t} - 0.2y^{\{1\}}_{t-1} - 0.1y^{\{1\}}_{t-2})$.

The second data is generated by a highly complex and hard to model piecewise-linear structure, and is given as
\begin{align*}
    y^{\{2\}}_{t} &= 30 + v^{\{2\}}_{t} \text{, if} \;\; y^{\{2\}}_{t-7} > 50 \text{,} \;\; y^{\{2\}}_{t-1} > 50 \text{,} \;\; \frac{1}{7}\sum_{k=1}^{7}y^{\{2\}}_{t-k} > 50 \;\; \\
    y^{\{2\}}_{t} &= 35 + v^{\{2\}}_{t} \text{, if} \;\; y^{\{2\}}_{t-7} > 50 \text{,} \;\; y^{\{2\}}_{t-1} > 50 \text{,} \;\; \frac{1}{7}\sum_{k=1}^{7}y^{\{2\}}_{t-k} < 50 \;\; \\
    y^{\{2\}}_{t} &= 40 + v^{\{2\}}_{t} \text{, if} \;\; y^{\{2\}}_{t-7} > 50 \text{,} \;\; y^{\{2\}}_{t-1} < 50 \text{,} \;\; \frac{1}{7}\sum_{k=1}^{7}y^{\{2\}}_{t-k} > 50 \;\; \\
    y^{\{2\}}_{t} &= 45 + v^{\{2\}}_{t} \text{, if} \;\; y^{\{2\}}_{t-7} > 50 \text{,} \;\; y^{\{2\}}_{t-1} < 50 \text{,} \;\; \frac{1}{7}\sum_{k=1}^{7}y^{\{2\}}_{t-k} < 50 \;\; \\
    y^{\{2\}}_{t} &= 56 + v^{\{2\}}_{t} \text{, if} \;\; y^{\{2\}}_{t-7} < 50 \text{,} \;\; y^{\{2\}}_{t-1} > 50 \text{,} \;\; \frac{1}{7}\sum_{k=1}^{7}y^{\{2\}}_{t-k} > 50 \;\; \\
    y^{\{2\}}_{t} &= 61 + v^{\{2\}}_{t} \text{, if} \;\; y^{\{2\}}_{t-7} < 50 \text{,} \;\; y^{\{2\}}_{t-1} > 50 \text{,} \;\; \frac{1}{7}\sum_{k=1}^{7}y^{\{2\}}_{t-k} < 50 \;\; \\
    y^{\{2\}}_{t} &= 66 + v^{\{2\}}_{t} \text{, if} \;\; y^{\{2\}}_{t-7} < 50 \text{,} \;\; y^{\{2\}}_{t-1} < 50 \text{,} \;\; \frac{1}{7}\sum_{k=1}^{7}y^{\{2\}}_{t-k} > 50 \;\; \\
    y^{\{2\}}_{t} &= 71 + v^{\{2\}}_{t} \text{, if} \;\; y^{\{2\}}_{t-7} < 50 \text{,} \;\; y^{\{2\}}_{t-1} < 50 \text{,} \;\; \frac{1}{7}\sum_{k=1}^{7}y^{\{2\}}_{t-k} < 50,
\end{align*}
where $v^{\{2\}}_{t}$ is a sample function from a stationary white Gaussian process with unit variance. 

For our synthetic data experiments, we combine $y^{\{1\}}_{t}$ and $y^{\{2\}}_{t}$ in different combinations to form different ensemble datasets. We create three different sets of data given as $y^{\{a\}}_{t}$, $y^{\{b\}}_{t}$ and $y^{\{c\}}_{t}$. These datasets are formed as:
\begin{align*}
    y^{\{a\}}_{t} &= 0.333y^{\{1\}}_{t} + 0.667y^{\{2\}}_{t} \text{, if} \;\; t\Mod{2}= 0 \;\; \\
    y^{\{a\}}_{t} &= 0.666y^{\{1\}}_{t} + 0.334y^{\{2\}}_{t} \text{, if} \;\; t\Mod{2}= 1,
\end{align*}
\begin{align*}
    y^{\{b\}}_{t} &= 0.200y^{\{1\}}_{t} + 0.800y^{\{2\}}_{t} \text{, if} \;\; t\Mod{4}= 0 \\
    y^{\{b\}}_{t} &= 0.400y^{\{1\}}_{t} + 0.600y^{\{2\}}_{t} \text{, if} \;\; t\Mod{4}= 1 \\
    y^{\{b\}}_{t} &= 0.600y^{\{1\}}_{t} + 0.400y^{\{2\}}_{t} \text{, if} \;\; t\Mod{4}= 2 \\
    y^{\{b\}}_{t} &= 0.800y^{\{1\}}_{t} + 0.200y^{\{2\}}_{t} \text{, if} \;\; t\Mod{4}= 3,
\end{align*}
\begin{align*}
    y^{\{c\}}_{t} &= 0.059y^{\{1\}}_{t} + 0.941y^{\{2\}}_{t} \text{, if} \;\; t\Mod{16}= 0 \\
    y^{\{c\}}_{t} &= 0.118y^{\{1\}}_{t} + 0.882y^{\{2\}}_{t} \text{, if} \;\; t\Mod{16}= 1 \\
    y^{\{c\}}_{t} &= 0.176y^{\{1\}}_{t} + 0.824y^{\{2\}}_{t} \text{, if} \;\; t\Mod{16}= 2 \\
    y^{\{c\}}_{t} &= 0.235y^{\{1\}}_{t} + 0.765y^{\{2\}}_{t} \text{, if} \;\; t\Mod{16}= 3 \\
    y^{\{c\}}_{t} &= 0.294y^{\{1\}}_{t} + 0.706y^{\{2\}}_{t} \text{, if} \;\; t\Mod{16}= 4 \\
    y^{\{c\}}_{t} &= 0.353y^{\{1\}}_{t} + 0.647y^{\{2\}}_{t} \text{, if} \;\; t\Mod{16}= 5 \\
    y^{\{c\}}_{t} &= 0.412y^{\{1\}}_{t} + 0.588y^{\{2\}}_{t} \text{, if} \;\; t\Mod{16}= 6 \\
    y^{\{c\}}_{t} &= 0.471y^{\{1\}}_{t} + 0.529y^{\{2\}}_{t} \text{, if} \;\; t\Mod{16}= 7 \\
    y^{\{c\}}_{t} &= 0.529y^{\{1\}}_{t} + 0.471y^{\{2\}}_{t} \text{, if} \;\; t\Mod{16}= 8 \\
    y^{\{c\}}_{t} &= 0.588y^{\{1\}}_{t} + 0.412y^{\{2\}}_{t} \text{, if} \;\; t\Mod{16}= 9 \\
    y^{\{c\}}_{t} &= 0.647y^{\{1\}}_{t} + 0.353y^{\{2\}}_{t} \text{, if} \;\; t\Mod{16}= 10 \\
    y^{\{c\}}_{t} &= 0.706y^{\{1\}}_{t} + 0.294y^{\{2\}}_{t} \text{, if} \;\; t\Mod{16}= 11 \\
    y^{\{c\}}_{t} &= 0.765y^{\{1\}}_{t} + 0.235y^{\{2\}}_{t} \text{, if} \;\; t\Mod{16}= 12 \\
    y^{\{c\}}_{t} &= 0.824y^{\{1\}}_{t} + 0.176y^{\{2\}}_{t} \text{, if} \;\; t\Mod{16}= 13 \\
    y^{\{c\}}_{t} &= 0.882y^{\{1\}}_{t} + 0.118y^{\{2\}}_{t} \text{, if} \;\; t\Mod{16}= 14 \\
    y^{\{c\}}_{t} &= 0.941y^{\{1\}}_{t} + 0.059y^{\{2\}}_{t} \text{, if} \;\; t\Mod{16}= 15.\\
\end{align*}

For all synthetic data experiments, our data samples are of length $730$, where the last $100$ samples are taken as the test data, and the remaining are taken as the training data. We evaluate the model accuracies on the test dataset. We train our ensemble models according to Fig. \ref{fig:alg_flow} and \textbf{Algorithm 1} and for the base models, we directly use $y^{\{1\}}_{t}$ and $y^{\{2\}}_{t}$. Therefore, we expect the ensemble models to learn the combination weights $w_{\boldsymbol{s}^E_{t}}^{(1)}$ and $w_{\boldsymbol{s}^E_{t}}^{(2)}$ used in forming all three ensemble datasets. As the side information vector $\boldsymbol{s}^E_{t}$, we provide the modulo information that splits the data weights into different regions. Note that the importance of the side information vector to the data as well as the data complexity varies among the three datasets, as there are different number of distinct weight combinations for each dataset.
For each of the three datasets, we fit our MLP ensemble model introduced in Section \ref{sec:mlp_ens} under all three constraints (unconstrained, affine constrained and convex constrained). Note that, as $w_{\boldsymbol{s}^E_{t}}^{(1)} + w_{\boldsymbol{s}^E_{t}}^{(2)} = 1$ is satisfied for all weight combinations, our ensemble algorithm should be theoretically able to capture all of the combination weights under any constraint.

\begin{figure}[!t]
    \centering
    \includegraphics[width=0.5\textwidth]{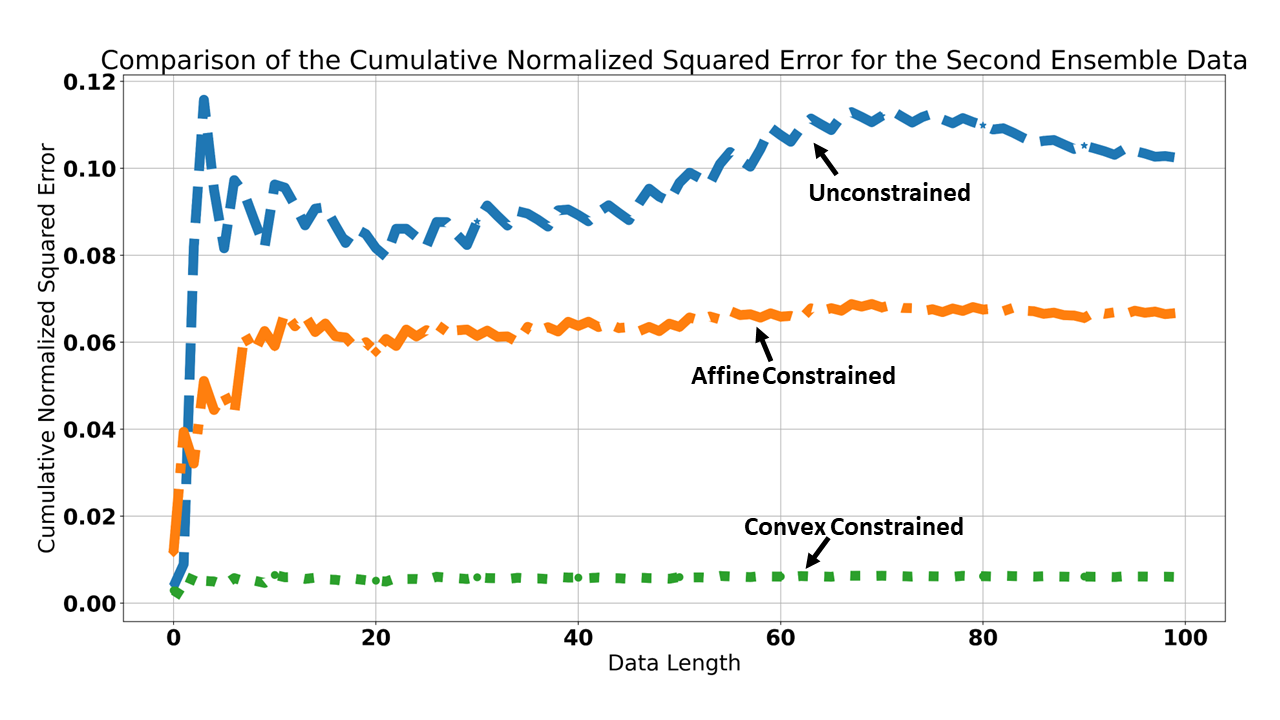}
    \caption{Comparison of the cumulative error for the prediction of the ensemble data $y^{\{b\}}_{t}$. The performance of the MLP ensemble model under unconstrained, affine constrained and convex constrained conditions are shown.}
    \label{fig:cumsum_error_b}
\end{figure}

\begin{figure}[!t]
    \centering
    \includegraphics[width=0.5\textwidth]{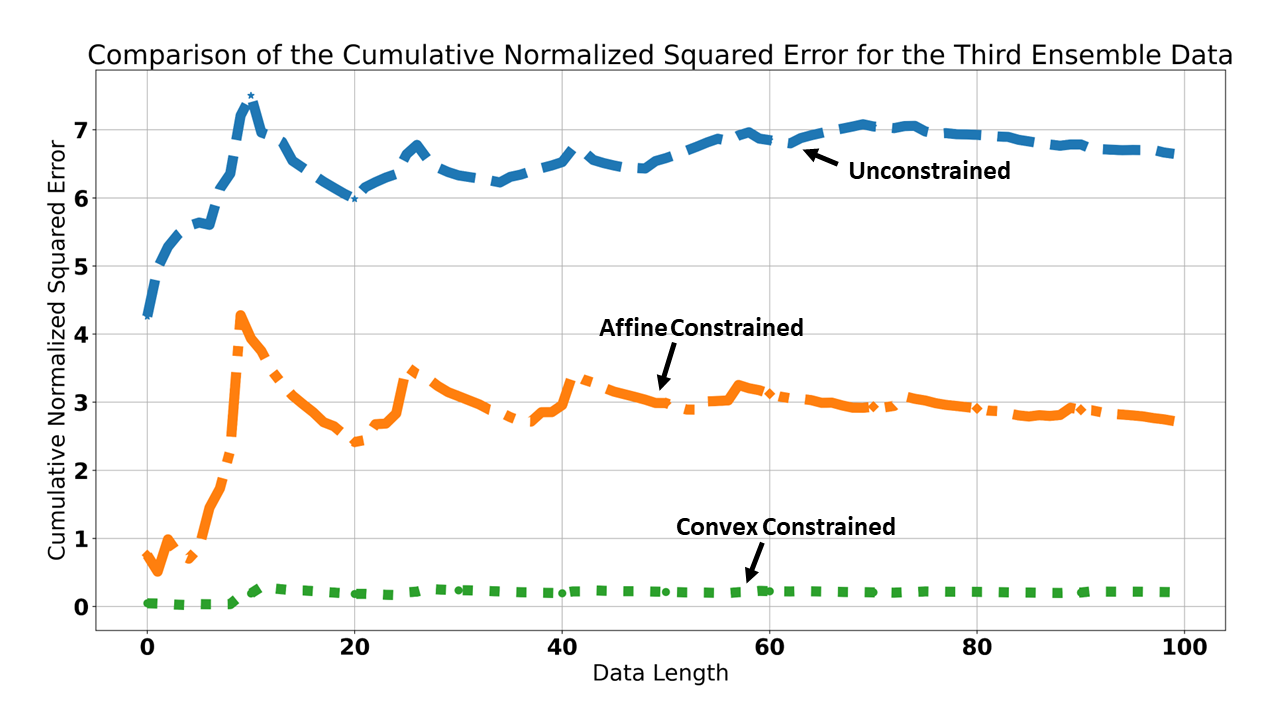}
    \caption{Comparison of the cumulative error for the prediction of the ensemble data $y^{\{c\}}_{t}$. The performance of the MLP ensemble model under unconstrained, affine constrained and convex constrained conditions are shown.}
    \label{fig:cumsum_error_c}
\end{figure}

Table \ref{table:synthetic} shows the results in terms of the final cumulative error for all three ensemble datasets, for all three constraints. We also illustrate the cumulative normalized total error for the predictions of $y^{\{b\}}_{t}$ and $y^{\{c\}}_{t}$ in Figs. \ref{fig:cumsum_error_b} and \ref{fig:cumsum_error_c}. When the data does not contain many distinct weight combinations related to the side information vector ($y^{\{a\}}_{t}$), all three ensembles are able to perfectly capture the weight combinations. However, as we increase the number of distinct weight combinations, the ensemble model under unconstrained and affine constrained conditions tend to make more errors compared to the convex constrained model. This is due to the fact that, these models have more parameters to learn compared to the convex constrained model, as the search space for the optimal combination weights is larger and therefore more complex. Therefore, the models may not converge to the optimal weight conditions, as in our example. Hence, even though the unconstrained ensemble model has the least expected error among all three constraints as explained in Section \ref{sec:theproposedmodel}, convex and affine models can be preferable in complex problems.

In addition, the MLP ensemble model under all three constraints outperform both the conventional linear ensemble and conventional MLP ensemble. In Table \ref{table:synthetic}, it is illustrated that both of the conventional methods are unable to capture the regional change in the structure of the observed sequence $y_{t}$, hence, producing large errors opposed to our MLP ensemble, which directly employs a side information vector containing the regional switching information. Thus, our results showcase the need of employing a data specific side information vector while learning the associated combination weights.

\begin{table}[!t]
    \begin{center}
        \begin{tabular}{ |c|c|c|c| }
            \hline
            Model & $y^{\{a\}}_{t}$ & $y^{\{b\}}_{t}$ & $y^{\{c\}}_{t}$ \\
            \hline
            \makecell{MLP Ensemble \\ Unconstrained} & 0.0 & 0.10248 & 6.64695 \\
            \hline
            \makecell{MLP Ensemble \\ Affine Constrained} & 0.0 & 0.06670 & 2.72202 \\
            \hline
            \makecell{MLP Ensemble \\ Convex Constrained} & 0.0 & $\boldsymbol{0.00603}$ & $\boldsymbol{0.21027}$ \\
            \hline
            \makecell{Conventional \\ Linear Ensemble} & 136.69053 & 256.78693 & 367.59445 \\
            \hline
            \makecell{Conventional \\ MLP Ensemble} & 124.90278 & 262.02210 & 140.68292 \\
            \hline
        \end{tabular}
    \end{center}
    \caption{Final cumulative error on the test set for all ensemble data, under all constraints for the MLP ensemble.}
    \label{table:synthetic}
\end{table}

\subsection{Total Residential Natural Gas Demand in Turkey}\label{sec:botas}
The data used in this section consists of the total residential natural gas demand in Turkey for 1000 days during the years 2018-2020, and each day is considered as a single sample. We train both our LightGBM ensemble introduced in Section \ref{sec:lgbm_ens} and MLP ensemble introduced in Section \ref{sec:mlp_ens}, under all weight constraints, according to Fig. \ref{fig:alg_flow} and \textbf{Algorithm 1}. For the base algorithms, we use the Seasonal Auto-Regressive Integrated Moving Average with eXogenous factors (SARIMAX), which is a linear model commonly used in time series forecasting \cite{box2015time}, and the LightGBM, which is a gradient boosting framework that uses tree based learning algorithms \cite{ke2017lightgbm}. We take the last 300 samples of the data as the test dataset, and the remaining as the training dataset. We evaluate the model accuracies on the test dataset.

\begin{figure*}[!h]
    \centering
    \includegraphics[width=0.65\textwidth]{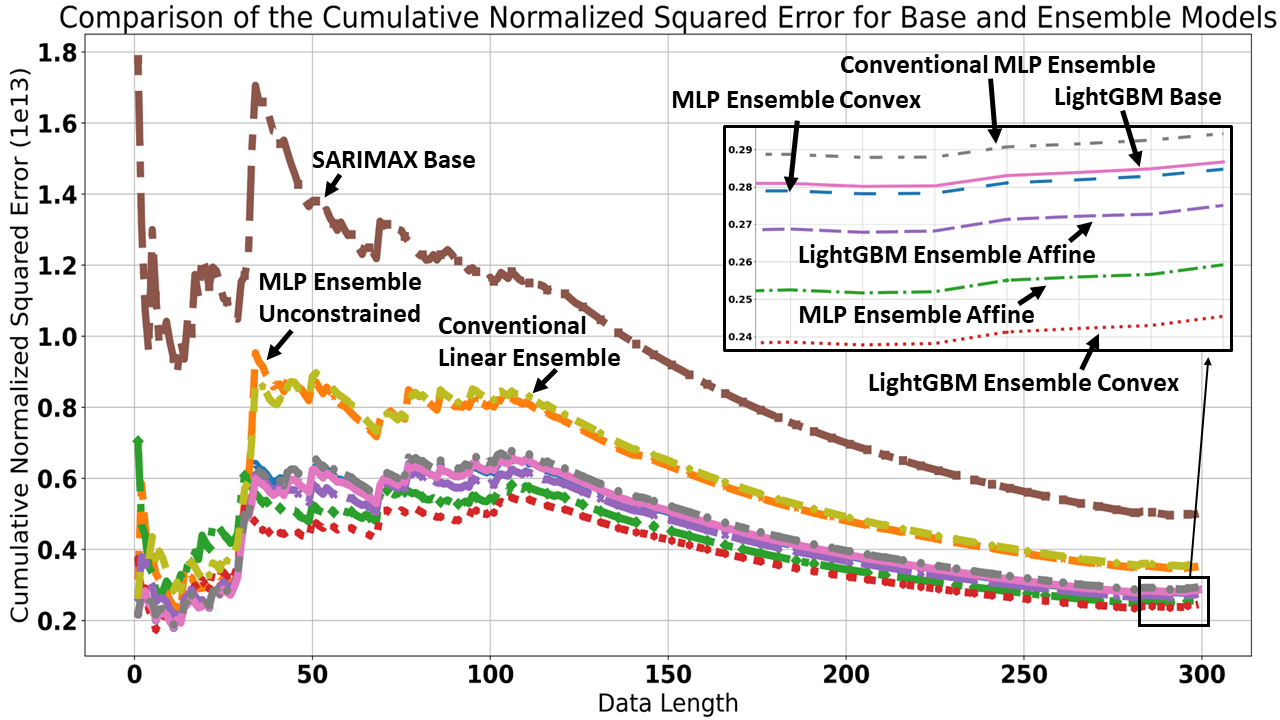}
    \caption{Comparison of the cumulative error for the prediction of the total residential natural gas demand in Turkey. The performance of our base and ensemble models under different constraints are shown. Note that the error plot for the unconstrained lightgbm ensemble is not provided here, as the errors produced by the model are high compared to other models.}
    \label{fig:cumsum_error_botas}
\end{figure*}

Table \ref{table:botas} illustrates our results in terms of the final cumulative error for both the base algorithms and the ensemble algorithms under all three constraints. We also illustrate the cumulative normalized total error for all models in Fig. \ref{fig:cumsum_error_botas}. Note that, the results for the LightGBM ensemble model under unconstrained weights condition is not shown, as the model could not converge to produce reasonable weights. In addition, both our results in Table \ref{table:botas} and Fig. \ref{fig:cumsum_error_botas} illustrate that both our LightGBM ensemble and MLP ensemble algorithms outperform the two base algorithms under both affine and convex weight constraints. For both ensemble models, when the weights produced by the models are unconstrained, the models could not improve the predictions produced by the base models. Therefore, as also illustrated in synthetic data experiments, even though the expected error for the unconstrained ensemble models are less than the ensemble models with constrained weights, convex and affine models can be preferable as they have less parameters to learn compared to the unconstrained case. In addition, our ensemble models under affine and convex constraints outperform the conventional ensemble methods, as illustrated in  \ref{table:botas} and Fig. \ref{fig:cumsum_error_botas}.

\begin{table}[!h]
    \begin{center}
        \begin{tabular}{ |c|c| } 
            \hline
            Model Name & Final Cumulative Error (1e10) \\
            \hline
            SARIMAX Base & 50.16 \\
            \hline
            LightGBM Base & 28.67 \\
            \hline
            \makecell{LightGBM Ensemble \\ Affine Constrained} & 27.51 \\
            \hline
           \makecell{LightGBM Ensemble \\ Convex Constrained} & $\boldsymbol{24.54}$ \\
            \hline
            MLP Ensemble Unconstrained & 35.25 \\
            \hline
            MLP Ensemble Affine Constrained & 25.92 \\
            \hline
            MLP Ensemble Convex Constrained & 28.47 \\
            \hline
            Conventional MLP Ensemble & 29.43 \\
            \hline
            Conventional Linear Ensemble & 35.86 \\
            \hline
        \end{tabular}
    \end{center}
    \caption{The final cumulative error for both the base and ensemble models used in predicting total residential natural gas demand in Turkey, under all constraints.}
    \label{table:botas}
\end{table}

\subsection{M5 Forecasting Dataset}

The M5 Forecasting dataset \cite{makridakis2022m5} involves the unit sales of the products sold by the retail company Walmart in USA. The products are classified in 3 different categories in 7 different departments, and are sold in 10 different stores in 3 different states. For our experiments, we examine the total number of unit sales in store CA\_3 department HOUSEHOLD\_1, therefore reducing the possibility of error in the data. The data length is 1941 days, where each day is considered a single sample. We consider the first 1841 days as the training dataset, and the last 100 days as the test dataset.

We train both our LightGBM ensemble introduced in Section \ref{sec:lgbm_ens} and MLP ensemble introduced in Section \ref{sec:mlp_ens}, under all weight constraints, according to Fig. \ref{fig:alg_flow} and \textbf{Algorithm 1}. For the base algorithms, we use the SARIMAX and LightGBM models, as in the case in Section \ref{sec:botas}. We evaluate both the base and ensemble model accuracies on the test dataset.

\begin{figure*}[!h]
    \centering
    \includegraphics[width=0.65\textwidth]{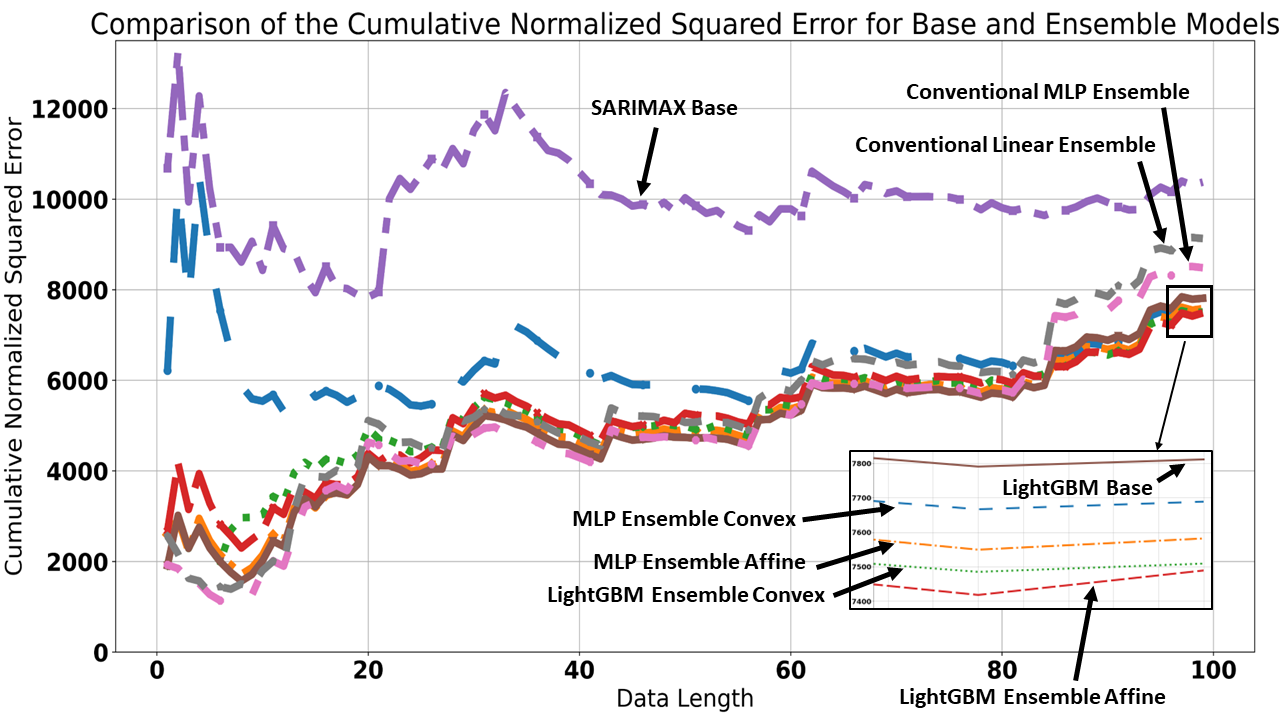}
    \caption{Comparison of the cumulative error for the prediction of the total unit sales in store CA\_3 department HOUSEHOLD\_1 of Walmart, USA. The performance of our base and ensemble models under different constraints are shown. Note that the error plots for the unconstrained LightGBM ensemble and unconstrained MLP ensemble are not provided here, as the errors produced by the model are high compared to other models.}
    \label{fig:cumsum_error_m5}
\end{figure*}

Table \ref{table:m5} illustrates our results in terms of the final cumulative error for both the base algorithms and the ensemble algorithms under all three constraints. We also give the cumulative normalized total error for all models in Fig. \ref{fig:cumsum_error_m5}. We do not show the results of the ensemble algorithms under unconstrained case, as the models could not converge to produce reasonable weights. Our results in Table \ref{table:m5} and Fig. \ref{fig:cumsum_error_m5} illustrate that both our MLP ensemble and LightGBM ensemble models outperform our base models and also the conventional ensemble models, under affine and convex constrained weights.

\begin{table}[!h]
    \begin{center}
        \begin{tabular}{ |c|c| } 
            \hline
            Model Name & Final Cumulative Error \\
            \hline
            SARIMAX Base & 10377.02 \\
            \hline
            LightGBM Base & 7811.49 \\
            \hline
            LightGBM Ensemble Affine Constrained & $\boldsymbol{7489.49}$ \\
            \hline
            LightGBM Ensemble Convex Constrained & 7509.36 \\
            \hline
            MLP Ensemble Affine Constrained & 7582.25 \\
            \hline
            MLP Ensemble Convex Constrained & 7688.94 \\
            \hline
            Conventional MLP Ensemble & 8485.79 \\
            \hline
            Conventional Linear Ensemble & 9130.51 \\
            \hline
        \end{tabular}
    \end{center}
    \caption{The final cumulative error for both the base and ensemble models used in predicting the total unit sales in store CA\_3 department HOUSEHOLD\_1 of Walmart, USA, under affine and convex constraints.}
    \label{table:m5}
\end{table}

\section{Conclusion}\label{sec:conclusion}
We studied the problem of predicting sequential time series data where we combine the predictions of multiple machine learning models using a novel ensembling approach. For the first time in the literature, we tackle the problem of finding the optimal combination weight vectors under unconstrained, affine constrained and convex constrained conditions, while considering a data specific side information vector. We analyzed the associated costs for learning all three constraints given the side information vector, and then introduced two novel and generic ensembling algorithms to find the optimal combination weight vectors under given constraints. In addition, we have presented a novel training scheme for ensemble models, where we have diminished any possible issues related to the training of the base models, such as co-linearity, high correlation and overfitting by the base algorithms. With various experiments containing synthetic and well-known real-life sequential data, we have illustrated superiority of our ensemble models over both the base models used in the experiments and the conventional ensembling methods in the literature.

\bibliographystyle{IEEEtran}
\bibliography{refs}

\end{document}